# A Survey of Large Language Models for Arabic Language and its Dialects

Malak O. Mashaabi

iWAN Research Group, College of Computer and Information Sciences, King Saud University, Riyadh, Saudi Arabia, malakmashabi@gmail.com

Shahad Z. Al- Khalifa

iWAN Research Group, College of Computer and Information Sciences, King Saud University, Riyadh, Saudi Arabia, shahadalkhalifa90@gmail.com

Hend S. Al-Khalifa*

iWAN Research Group, College of Computer and Information Sciences, King Saud University, Riyadh, Saudi Arabia, hendk@ksu.edu.sa

**Abstract.** This survey offers a comprehensive overview of Large Language Models (LLMs) designed for Arabic language and its dialects. It covers key architectures, including encoder-only, decoder-only, and encoder-decoder models, along with the datasets used for pre-training, spanning Classical Arabic, Modern Standard Arabic, and Dialectal Arabic. The study also explores monolingual, bilingual, and multilingual LLMs, analyzing their architectures and performance across downstream tasks, such as sentiment analysis, named entity recognition, and question answering. Furthermore, it assesses the openness of Arabic LLMs based on factors, such as source code availability, training data, model weights, and documentation. The survey highlights the need for more diverse dialectal datasets and attributes the importance of openness for research reproducibility and transparency. It concludes by identifying key challenges and opportunities for future research and stressing the need for more inclusive and representative models.

## 1 INTRODUCTION

Large-language models (LLMs) have gained significant attention owing to their high performance in a wide range of natural language tasks. These models learn to understand and generate language by training billions of parameters on vast volumes of textual data. Despite being a relatively new field, LLM research is advancing rapidly in various directions, including the development of models for different languages and dialects.

Arabic, the fifth most spoken language in the world, presents unique challenges and opportunities in the field of natural language processing (NLP). The Arabic language is characterized by its diglossia, where written formal Arabic (Classical Arabic and Modern Standard Arabic) differs significantly from spoken vernacular varieties (Dialectal Arabic). This linguistic phenomenon, along with the substantial dialectal variation and morphological complexity of Arabic, poses challenges for language modeling.

* Corresponding author

Early Arabic language models relied on statistical approaches and primarily focused on Modern Standard Arabic (MSA). These models were trained on relatively small datasets and were limited in their ability to capture the complexity and diversity of the Arabic language [1]. However, recent advancements in Arabic LLMs have been driven by the adoption of transformer-based architectures and availability of large-scale pretraining datasets. Models such as AraBERT [2], QARiB [3], and MARBERT [4] have achieved state-of-the-art performance in various Arabic NLP tasks, including sentiment analysis, named entity recognition, and question answering. These models were trained on diverse datasets, including web-crawled data, news articles, and social media, enabling them to capture the nuances of MSA, Dialectal Arabic (DA), and Classical Arabic (CA), which hold significant importance in formal and religious contexts, but may still be underrepresented in modern language models.

This survey aims to provide a comprehensive overview of the current state of LLMs in Arabic NLP. It focuses on various architectures used for Arabic LLMs, including encoder-only models (for example, BERT [5], ELECTRA [6] and RoBERTa [7]), decoder-only (for example, GPT-2 [8]), and encoder-decoder models (for example, T5 [9]). The survey examined the training datasets used for these models, which encompass diverse sources, such as web crawled data, news articles, books, and social media. It also discusses the evaluation metrics employed to assess the performance of Arabic LLMs in downstream tasks such as sentiment analysis, named entity recognition, and question answering.

Additionally, the survey examined the openness of Arabic LLMs, considering factors such as the availability of source code, training data, model weights, and documentation. The openness of these models is crucial for reproducibility and transparency and facilitates further research and development in the field of Arabic NLP. The objectives of this survey are as follows:

1. To provide a comprehensive overview of the current state of LLMs in Arabic NLP, focusing on various architectures, training datasets, downstream tasks and evaluation metrics.
2. To explore the datasets used for pretraining Arabic LLMs, covering CA, MSA, DA, and combinations of these Arabic forms.
3. To assess the openness of Arabic LLMs based on the availability of source code, training data, model weights, and documentation.

By addressing these objectives, this survey aims to serve as a valuable resource for researchers, practitioners, and enthusiasts interested in advancing Arabic NLP through large language models. It provides insights into the current landscape of Arabic LLMs, their strengths and limitations, and the potential for future development in this rapidly evolving field.

The rest of this paper is structured as follows: Section 2 presents the methodology used for gathering research papers for our survey on LLMs for the Arabic language and its dialects. Section 3 explores the various architectures employed by Arabic language models, focusing on their designs and functionalities. Section 4 provides an overview of the datasets and resources used for training Arabic LLMs categorized by Arabic form, including CA, MSA, DA, and combinations of different Arabic forms. Section 5 delves into the realm of monolingual, bilingual, and multilingual LLMs for the Arabic language and its dialects, discussing their architecture and the array of downstream tasks they are designed to tackle. Section 6 assesses the openness of Arabic LLMs based on a comprehensive framework considering factors such as the availability of source code, training data, model weights, and documentation. Section 7 concludes the survey by summarizing the key findings and highlighting the potential for further advancements in Arabic LLMs.

## 2 METHOD

In this section, we detail our method for gathering research papers for our survey on LLMs for the Arabic language and its dialects. Our process consisted of three main stages: keyword-based search and data collection, data verification, and data analysis. Initially, we collected 80 research papers through our keyword-based search. After the verification and analysis stages, we refined this set to 34 papers. The overall process is illustrated in Figure 1. In the following subsections, we describe these three stages in detail.

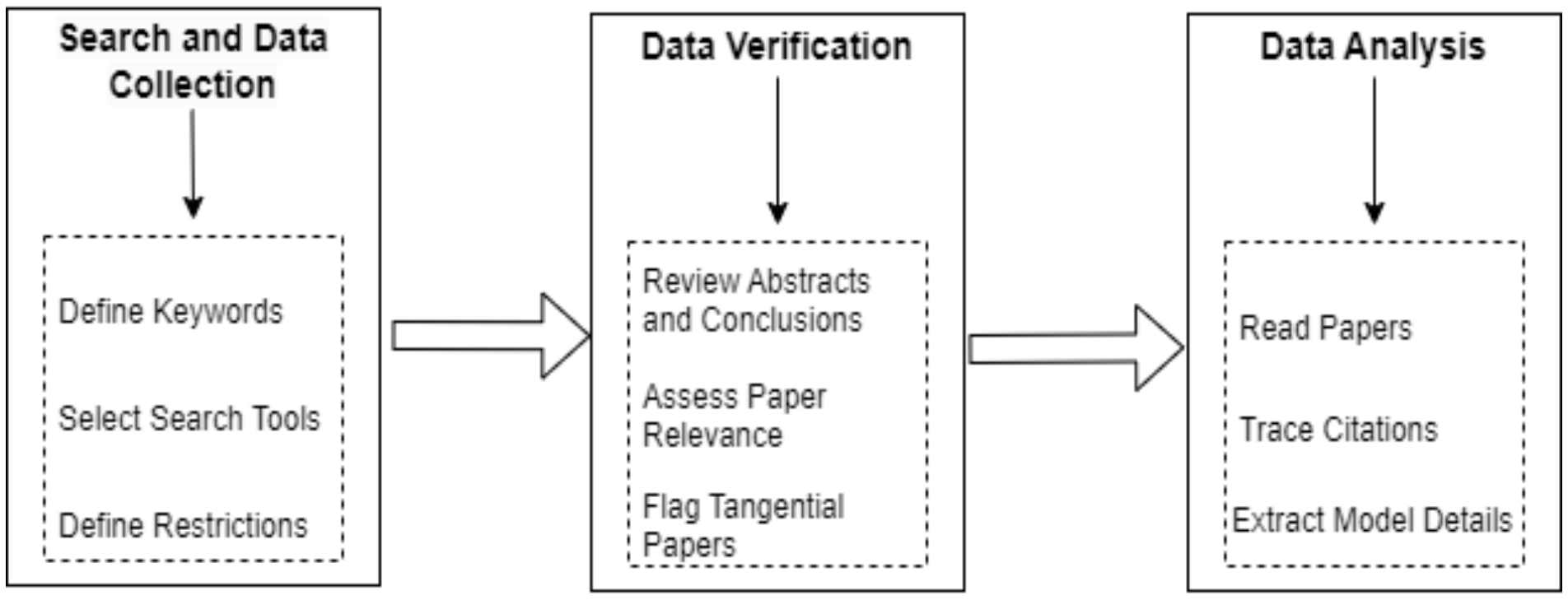

Figure 1: Process for Gathering and Analyzing Research Papers on Arabic LLMs and its dialects

### 2.1 Stage One: Search and Data Collection

In this stage, we aimed to gather a comprehensive collection of research papers relevant to LLMs for the Arabic language and its dialects. Our initial search query was customized to capture a broad range of Arabic LLMs by using keywords such as "Arabic LLM", "Arabic large language model", and "Arabic dialect LLM". These keywords were chosen to ensure the inclusion of both general LLMs, focusing on different forms of Arabic, and those addressing regional dialects. We restricted our search to LLMs that have a technical paper associated with them, excluding models that are mentioned only on platforms such as HuggingFace or in news without a published paper, these include: Noor [10], Noon [11], SILMA [12], Mulhem [13], Huawei's Arabic LLM [14], and Mistral Saba[1]. The time frame for collecting papers was up to February 2025 to capture the most recent advancements in Arabic LLMs.

The data collection process utilizes widely used academic search engines, such as Google Scholar and Semantic Scholar. In addition, we conducted a supplementary search using Litmaps[2], which tracks citation networks, allowing us to further expand our dataset. By outlining the connections between the cited works in each paper, we minimize the possibility of missing relevant studies.

### 2.2 Stage Two: Data Verification

To ensure the relevance and quality of the collected papers, we implemented a verification process designed to maintain the integrity of our survey. Each paper's abstract and conclusion were carefully reviewed to assess whether the research aligned with our specific focus on Arabic LLMs and dialectal variation. Papers that addressed related but tangential topics, such as models fine-tuned for specific NLP tasks or multimodal LLMs, were flagged for further review or exclusion.

1 https://mistral.ai/news/mistral-saba
2 https://www.litmaps.com/

Through this verification process, we selected a refined and high-quality collection of research papers that directly contributed to our survey on Arabic LLMs. This ensured that our analysis was based on the most relevant and impactful work in this field.

### 2.3 Stage Three: Data Analysis

Our analysis process was designed to ensure a thorough understanding of the Arabic LLMs. The process involved the following steps:

1. Each paper was carefully read and analyzed to extract relevant information related to Arabic LLMs. We focused on understanding the methodologies, data sources, and results of each study.
2. To ensure that no relevant papers were missed, we identified additional Arabic LLMs referenced in the related work sections of the papers that we initially collected. By tracing citations and references, we discovered and included further studies that were not part of our original search results, thereby broadening the scope of our survey and ensuring a more complete representation of the topic.
3. For each identified Arabic LLM, detailed information was extracted, including the model's architecture, dataset used for training, tasks it was applied to, and evaluation metrics reported. This step was crucial for comparing different models and identifying trends and patterns in Arabic LLMs research.

## 3 ARABIC LARGE LANGUAGE MODELS ARCHITECTURES

This section explores the various architectures employed by Arabic language models, focusing on their designs and functionalities. The architectures of these models typically fall into three main categories: encoder, decoder, and encoder-decoder models. Each of these architectures plays a unique role in NLP tasks, affecting how models understand and generate Arabic text. Table 1 summarizes encoder-only models, while Table 2 summarizes decoder-only and encoder-decoder models. This overview highlights the diversity in model design, reflecting specific use cases and strengths across different NLP applications.

### 3.1 Encoder-only Models

These models only use an encoder network. They were first developed for tasks that require understanding text, such as predicting categories for input text in text classification and next sentence prediction [5]. Notable examples include Bidirectional Encoder Representations from Transformers (BERT) and its versions, ELECTRA and RoBERTa, as described below:

1. BERT [5] is one of the most prevalent encoder-transformer models and operates bidirectionally, meaning that it is trained to learn the context of words in a sentence by examining both preceding and subsequent words. The BERT model architecture comprises three components:
    a. An embedding module that generates a series of embedding vectors from the input text.
    b. A stack of Transformer encoders that produce contextual representation vectors from these embedding vectors.
    c. A fully connected layer that transforms the representation vectors into one-hot vectors.

In addition, BERT is trained with two primary objectives:

a. Masked Language Modeling (MLM), which involves predicting missing words in a sentence based on the surrounding context.

b. Next Sentence Prediction (NSP), which aims to determine whether one sentence logically follows another.

The success of BERT has led to the development of several Arabic models such as AraBERT [2] and MARBERT [4].

2. ELECTRA [6] is a pre-training method designed as an alternative to BERT, focusing on efficiency and improving performance in NLP tasks. Instead of using MLM like BERT, which predicts masked words in a sentence, ELECTRA employs a method called "replaced token detection." This approach replaces some tokens in the input text with plausible alternatives that are generated by a small generator network. The main objective of the model is to act as a discriminator to identify whether each token is original or replaced. The ELECTRA model consists of two parts:
    a. A generator, which creates plausible alternative tokens for the input.
    b. A discriminator that predicts whether the tokens in the input have been replaced

This method is more sample-efficient than MLM because it evaluates all tokens rather than just the small subset masked in BERT. An example of an Arabic model that uses this approach is AraELECTRA [15].

3. RoBERTa [7], which stands for Robustly Optimized BERT Pretraining Approach, is an enhanced version of BERT that focuses on improving the pretraining process to achieve better performance. Several Arabic models such as AraRoBERTa [16] have adopted this approach. The key improvements of RoBERTa over BERT include the following:
    a. RoBERTa is trained for a longer period compared to BERT, using more data and larger mini-batches, which helps the model to learn better representations.
    b. Unlike BERT, RoBERTa eliminates the NSP objective that was originally designed to predict whether two sentences are consecutive in a text. The removal of the NSP improved downstream performance in multiple tasks.
    c. RoBERTa processes longer sequences of text during training, allowing it to learn from more contextual information at once.
    d. Instead of statically masking tokens once during preprocessing, RoBERTa applies dynamic masking, where the mask pattern changes during training to improve the model's ability to generalize.

Table 1: Overview of Encoder-Only Architecture in Arabic LLMs

| Model Name | Parameters | Number of Tokens | Original Source for training data | Derived Arabic LLMs | Release date |
|---|---|---|---|---|---|
| BERT | 110, 340M | 137B tokens | en-Wiki, Books | QARiB [3] | 2020 |
| | | | | AraBERT [2] | |
| | | | | GigaBERT [17] | |
| | | | | ArabicBERT [18] | |
| | | | | CAMeLBERT [19] | 2021 |
| | | | | ARBERT [4] | |
| | | | | MARBERT [4] | |
| | | | | SudaBERT [20] | |
| | | | | AraLegal-BERT [21] | 2022 |
| | | | | JABER [22] | |
| | | | | SABER [22] | |
| | | | | DziriBERT [23] | |
| | | | | AraBART [24] | |
| | | | | TunBERT [25] | 2023 |

| Model Name | Parameters | Number of Tokens | Original Source for training data | Derived Arabic LLMs | Release date |
|---|---|---|---|---|---|
| | | | | MorrBERT [26]<br>SaudiBERT [27]<br>AlcLaM [28]<br>EgyBERT [29]<br>AraPoemBERT [30]<br>DarijaBERT [31] | 2024 |
| ELECTRA | 110M | | en-Wiki, Books | AraELECTRA [15] | 2021 |
| RoBERTa | 355M | 2.2T tokens | en-Wiki, Books, Reddit, CC-news &stories | AraRoBERTa [16]<br>MorRoBERTa [26] | 2022<br>2023 |

### 3.2 Decoder-only Models

Decoder models are designed to produce or recreate output data from internal representations, turning abstract information into a more accessible format [32]. They are particularly effective in translating complex data structures, such as latent features or encoded information, into coherent outputs. These models are frequently used in tasks such as natural language generation, image and video creation, and conversion of encoded data into readable formats. They play a crucial role in language modeling, content creation, and machine translation, where sequential logical output is necessary. Models like Generative Pretrained Transformer (GPT) represent decoder-only architectures that generate text using a unidirectional method, predicting the next word in a sequence by focusing solely on the preceding context. This makes them particularly suited for text creation and language processing tasks [33]. Notable examples include GPT2, GPT3, LlaMA, OLMo, Gemma, and Stable LM 2 as described below:

1. GPT-2 [8] was a pioneer in creative text generation, utilizing 1.5 billion parameters trained on a 40GB text dataset. It was built upon the foundational architecture introduced by GPT-1, which was the first powerful decoder-only transformer model developed by OpenAI. GPT-1 demonstrated that unsupervised pretraining combined with task-specific fine-tuning could achieve state-of-the-art (SOTA) performance but displayed some limitations such as difficulty with long contexts and occasional nonsensical outputs. GPT-2 scaled this approach, vastly increasing the number of parameters and improving its ability to generate coherent and creative text over longer sequences. For example, AraGPT2 [34] applied this architecture to Arabic text generation.
2. GPT-3 takes the concept even further by expanding the model to 175 billion parameters, showing that performance improves with scale and can compete with fine-tuned models. GPT-3's larger scale allows it to handle not only creative text generation but also more complex tasks, such as question answering (QA) and machine translation (MT). Despite its vast capabilities, GPT-3, similarly to GPT-2, remains exposed to biases present in its training data. JASMINE [35] and Jais [36] are examples of models that follow GPT-3 architecture.
3. The LLaMA [37] series of decoder-only LLMs, with models ranging from seven to 70 billion parameters, is well regarded for its parameter efficiency and instruction-tuning capabilities. LLaMA-2 [38] focuses on fine-tuning the LLaMA-2-Chat model for safer and more effective dialogue generation. This model benefits from 40% more training data, extended context length, and the use of grouped-query attention to boost performance. AceGPT [39] follows the LLaMA-2 architecture and is optimized for Arabic text generation and dialogue systems.
4. OLMo [40] (Open Language Model) is an open-source, decoder-only language model developed by the Allen Institute for Artificial Intelligence (AI2). Unlike many proprietary models, OLMo prioritizes openness by providing full training data, training and evaluation code, intermediate model checkpoints, and training logs. The architecture of OLMo follows modern Transformer-based decoder models, incorporating SwiGLU activation,

RoPE positional embeddings, and non-parametric layer normalization to improve training stability and efficiency. The model was trained on over 2 trillion tokens from an open-accessible dataset. Fanar [41] is an example of an Arabic LLM that follows the OLMo architecture for Arabic text understanding and generation.

5. Gemma 2 [42] developed by Google DeepMind and other Google teams, is one of the latest models in the Gemma family of lightweight, state-of-the-art open models. It offers parameter sizes ranging from 2 billion to 27 billion and features several key enhancements, including interleaving local-global attentions and adopting group-query attention for improved efficiency. Gemma 2 also utilizes knowledge distillation in its 2B and 9B models, allowing it to outperform models that are twice their size. Atlas-Chat [43] is an example of an Arabic LLM that follows the Gemma 2 architecture for Arabic text understanding and generation.
6. Stable LM 2 [44] is an open-source decoder-only Transformer developed by Stability AI. It follows an autoregressive sequence modeling approach and features a context length of 4096 tokens. The model is optimized for efficiency, utilizing FlashAttention-2 for improved sequence-wise parallelism. The training dataset consists of approximately 2 trillion tokens, including multilingual and diverse domain sources such as RefinedWeb, The Pile, RedPajama, and OSCAR. Stable LM 2 is designed for multilingual capabilities, including English, Spanish, German, Italian, French, Portuguese, and Dutch. Arabic Stable LM [45] is an example of a model derived from the Stable LM 2 architecture, fine-tuned specifically for Arabic text understanding and generation.

### 3.3 Encoder-decoder Models

Encoder-decoder models are well-suited for tasks that require understanding and generating responses to the input text. A prominent example of this architecture is the Text-to-Text Transfer Transformer (T5) [9] that employs a unified text-to-text training framework for a wide range of NLP tasks. It implements a unique approach to MLM by masking the entire length of consecutive words rather than single tokens, which accelerates training by dealing with shorter sequences. Additionally, T5 optimizes its performance using adapter modules that fine-tune the model for specific NLP tasks during the fine-tuning phase. This allows the model to adjust efficiently to task-specific requirements without the need to retrain the entire model. AraMUS [46] and AraT5 [47] are examples of Arabic models that are based on this architecture.

Table 2: Overview of Decoder-Only and Encoder-Decoder Architectures in Arabic LLMs

| Architecture Type | Model Name | Parameters | Number of Tokens | Original Source for training data | Derived Arabic LLMs | Release date |
|---|---|---|---|---|---|---|
| Decoder-only Models | GPT-2 | 1.5B | 10B tokens | Reddit | AraGPT2 [34]<br>AraQA [48]<br>ArabianGPT [49] | 2021<br>2023<br>2024 |
| | GPT-3 | 175B | 300B tokens | Wikipedia, CC 2016-19, Books, Web text | JASMINE [35]<br>Jais and Jais-chat [36] | 2023 |
| | LLaMA2 | 7,13,34,70B | 2T tokens | Web data | AceGPT [39]<br>AraStories [50]<br>ALLaM [51]<br>AraLLaMA [52] | 2024<br><br><br>2024<br>2025 |
| | OLMo | 1, 7B | 2T, 2.46T tokens | Common Craw, GitHub, Reddit, Semantic Scholar, Project Gutenberg, Wikipedia | Fanar Star [41] | |

| Architecture Type | Model Name | Parameters | Number of Tokens | Original Source for training data | Derived Arabic LLMs | Release date |
|---|---|---|---|---|---|---|
| | Gemma 2 | 2,9,27B | 2T, 8T, 13T tokens | Web documents, Code, Scientific articles | Atlas-Chat [43]<br>Fanar Prime [41] | 2024<br>2025 |
| | Stable LM | 1.6B | ~ 2T tokens | Academic, Book, Web, Social, Law, Math, Wikipedia, code and Instruction | Arabic Stable [45] | 2024 |
| Encoder-decoder Models | T5 | 223M | 156B tokens | CommonCrawl | AraT5 [47]<br>AraMUS [46] | 2022<br>2023 |

## 4 DATASETS FOR ARABIC LLMS PRETRAINING

LLMs are typically pre-trained on vast amounts of text from the Internet, including websites, books, and articles. The quality and variety of these data directly affect the performance and capabilities of LLMs. This section provides an overview of the resources used to train Arabic LLMs.

### 4.1 Overview of Arabic Language Datasets

Arabic datasets for the development of advanced LLMs are growing rapidly, offering increasingly diverse datasets. These datasets help in improving natural language processing tasks in the Arabic language domain. In addition, they play a key role in enhancing the quality of LLMs in understanding and generating Arabic text. This will contribute to better communication technologies and digital solutions for Arabic-speaking communities and beyond. In the following sections, we provide details of the resources used for Arabic LLMs pre-training, categorized by Arabic forms. We begin with the CA form, followed by the MSA form, the DA form, and finally the combinations of different Arabic forms of resources.

### 4.2 Classical Arabic (CA) Resources

Classical Arabic (CA) is the language of the Quran and a standardized literary form of Arabic that has been used since the 7th century [33]. Despite its historical nature, CA remains relevant for various applications including religious studies, historical research, and cultural preservation. CA is primarily used in religious and historical contexts and plays a significant role in developing specialized Arabic LLMs despite its limited use in everyday life. The limited availability of digitized CA compared with MSA and DA presents challenges for LLMs training. However, projects such as the OpenITI corpus have made significant contributions to this field. The latest OpenITI[3] release includes 10,202 text files representing 6,236 unique works by 2,582 authors, focusing primarily on pre-modern Arabic literature. Table 3 provides an overview of the models pre-trained on CA resources. Although the number of models focusing on CA is currently limited, the sizeable dataset used for CAMeLBERT-CA demonstrates the potential for developing more LLMs for this Arabic form.

Table 3: CA Datasets for Arabic LLMs. Data as of February 2025.

| Model Name | Dataset Size | Tokens/Number of words | Source |
|---|---|---|---|
| CAMeLBERT-CA [19] | 6GB | 847M words | OpenITI corpus (v1.2) |

[3] https://openiti.org/projects/OpenITI%20Corpus.html

## 4.3 Modern Standard Arabic (MSA) Resources

Modern Standard Arabic (MSA) is the standardized and literary form of Arabic used in writing and formal speech throughout the Arab world that developed in the late 19th and early 20th centuries [53]. It serves as the foundation for most Arabic NLP tasks, such as news analysis and official document processing. This is critical for LLMs development due to its formal and standardized nature. MSA's importance derives from its role as a formal language in the Arab world, bridging the gap between various Arabic dialects and serving as the primary language for written media, literature, and formal communication. Recent years have seen significant growth in MSA resources, reflecting the increasing demand for Arabic language technologies. These datasets typically consist of web-crawled data, news articles, Wikipedia content, and other formal Arabic texts. Table 4 provides an overview of the models pre-trained on MSA resources.

Table 4: MSA Datasets for Arabic LLMs. Data as of February 2025.

| Model Name | Dataset Size | Tokens/Number of words | Data Source |
|---|---|---|---|
| AraBERT [2] | ~24 GB | 64K tokens | Arabic news websites, 1.5 billion words Arabic corpus, and OSIAN Corpus |
| GigaBERT-v0 [17] | Not explicitly mentioned | 4.7B tokens (3.6B for English and 1.1B for Arabic) | Gigaword corpus |
| GigaBERT-v1 [17] | Not explicitly mentioned | 7.4B tokens (6.1B for English and 1.3B for Arabic) | Gigaword and Wiki |
| GigaBERT-v2/3/4 [17] | Not explicitly mentioned | 10.4B tokens (6.1B for English and for 4.3B Arabic) | Gigaword, Wiki, and Oscar |
| CAMeLBERT-MSA [19] | 107GB | 12.6 billion words | Arabic Gigaword Fifth Edition, Abu El-Khair Corpus, OSIAN corpus, Arabic Wikipedia 2019, and unshuffled version of the Arabic OSCAR corpus |
| AraELECTRA [15] | 77 GB | 8.8 billion words | OSCAR corpus, 1.5 billion words Arabic Corpus, Arabic Wikipedia dump from September 2020, OSIAN corpus, and news articles provided by As-Safir newspaper |
| AraGPT2 [34] | 77 GB | ~ 9.93 billion words | The unshuffled OSCAR corpus, Arabic Wikipedia dump, 1.5 billion words Arabic Corpus, OSIAN corpus, and news articles provided by As-safir newspaper |
| AraBART [24] | 73 GB | 50K tokens | Same one used to pretrain AraBERT, without preprocessing. |
| AraLegal-BERT [21] | 4.5 GB | ~ 64K words | Books (Master and PhD theses, research papers, magazines, dictionaries and Fiqh books), legal cases (Legal Cases in KSA and Gulf countries which consists of copy rights, design rights, facts and appealing), terms and laws (Laws and regulations in KSA and Gulf countries), and reports and studies (Reports and studies, academic courses, forms, reports, contracts) |
| JABER [22] | 115 GB | 64k vocabulary size | Arabic Wikipedia, EL-KHEIR corpus, OSIAN corpus, and Common Crawl |
| SABER [22] | 115 GB | 64k vocabulary size | Arabic Wikipedia, EL-KHEIR corpus, OSIAN corpus, and Common Crawl |
| Jais and Jais-chat [36] | Not explicitly mentioned | 395 billion tokens (116B tokens for Arabic, 232B | Abu El-Khair, Aranews, ArabicText 2022, AraC4, Arabic Wikipedia, ArabicNews 2020, Maktabah, UN Meeting transcripts, and Other Sources. Pile-CC (subset of Common Crawl), Books3, ArXiv, PubMed Central, OpenWebText2, Wikipedia (English), FreeLaw, PubMed Abstracts, DeepMind |

| Model Name | Dataset Size | Tokens/Number of words | Data Source |
|---|---|---|---|
| | | tokens for English, and 46B tokens for codes) | Mathematics, Project Gutenberg (PG-19), BookCorpus2, EuroParl, PhilPapers, YouTube Subtitles, NIH Grant Abstracts, and Enron Emails as an English resource.<br>GitHub as code resource |
| AraQA [48] | 27,778 question-answer pairs | Not explicitly mentioned | Scraped from trusted websites on the internet containing Islamic question-answer pairs |
| ArabianGPT-0.1B [49] | 15.5 GB | 1.8 billion tokens | Arabic newspaper articles |
| ArabianGPT-0.3B [49] | 23 GB | ~ 3.35 billion tokens. | Composite Linguistic Resource |
| AceGPT-7B [39] | Not explicitly mentioned | 30 billion tokens (19.2B tokens for Arabic and 10.8B tokens for English) | Arabic text 2022, Arabic Wikipedia, CC100, and OSCAR3.<br>Slim Pajama as an English resource. |
| AceGPT-13B [39] | | 10 billion tokens (6B tokens for Arabic and 4B tokens for English) | Arabic text 2022, Arabic Wikipedia, CC100, and OSCAR3.<br>Slim Pajama as an English resource. |
| ALLaM [51] | Not explicitly mentioned | 540 billion tokens for Arabic and 4 trillion tokens for English | Web documents, news articles, books (literature, religion, law and culture, among others), Wikipedia, and audio transcripts (books and news).<br>Dolmav1 and Pile as an English resources |
| Arabic Stable [45] | Not explicitly mentioned | 734 billion tokens (619B tokens for English and 115B tokens for Arabic) | CulturaX, SANAD, Arabic E-book corpus, and English texts |
| AraLLaMA [52] | Not explicitly mentioned | ~ 115.3 billin tokens for Arabic | Common Crawl, WebText, Books, Newspapers, Wikipedia1, and Wikipedia2.<br>Slim Pajama and Proof-Pile-2 as an English resource. |

The available datasets used to pre-train MSA language models are as follows:

1. **1.5 billion Words Arabic Corpus[4]:** It is also called Abu El-Khair Corpus or El-Khair Corpus. The corpus contains over 1.5 billion words extracted from more than five million newspaper articles collected between 2002 and 2014. The data comes from ten major news platforms across eight Arab countries: UAE, Algeria, Saudi Arabia, Syria, Kuwait, Lebanon, Egypt, and Yemen.

[4] https://arxiv.org/pdf/1611.04033v1

2. **OSIAN Corpus[5]:** The Open-Source International Arabic News (OSIAN) Corpus contains 3.5 million articles with over 37 million sentences and approximately one billion tokens. It was gathered from various Arabic news sources including CNN, DW, RT, and Aljazeera during a web crawl conducted in March 2018.
3. **Gigaword Corpus[6]**: It is introduced by the Linguistic Data Consortium (LDC) in 2003, The corpus consists of Arabic newswire from four sources: Agence France Presse, Al Hayat, Al Nahar, and Xinhua. It contains 319 files, totaling 4,348 MB and 391,619 words. There are multiple editions of the Gigaword Corpus, including the Arabic Gigaword Second Edition, Third Edition, Fourth Edition, and the final release, which is the Fifth Edition. The Arabic Gigaword Fifth Edition includes all the content from the previous editions and new data covering the period from January 2009 through December 2010.
4. **Oscar 2019 Corpus[7]:** It is the initial release of the Open Super-large Crawled Aggregated Corpus (OSCAR). This open-source project was designed to provide extensive multilingual datasets from the Common Crawl dataset. The Arabic version contains 3.17 billion unique words and is 32 GB in size. Later versions include OSCAR 21.09, OSCAR 22.01, and OSCAR 23.01.
5. **Arabic Wikipedia Dump[8]:** The Arabic Wikipedia dataset is generated by the Wikimedia Foundation and consists of a full database dump of the Arabic Wikipedia at specific points in time. For example, the Arabic Wikipedia Dump 2019 was created on February 1, 2019.
6. **ArabicText 2022[9]**: It is a large collection of publicly available Arabic web data, totaling over 200GB in size. It includes sources like ArabicWeb22-A, ArabicWeb16, OSCAR3, ArabicWeb22-B, CC100, Abu El-Khair Corpus, Arabic Tweets, and Arabic Wikipedia.
7. **AraC4[10]:** It is a part of the Multilingual Colossal Clean Crawled Corpus (mC4), this dataset extends the C4 corpus to 101 languages, including Arabic. The Arabic subset contains 92,547,752 examples, derived from Common Crawl web crawls. However, the exact date range for the Arabic subset is not explicitly documented.
8. **Maktabah[11]:** contains over 6,500 Arabic books, divided into more than six million paragraphs. The exact date range of the collected data is not explicitly documented.
9. **TyDi dataset[12]:** It is a question-answer dataset with 204K question-answer pairs, , covering 11 typologically diverse languages. The dataset was introduced in 2020, with its source data collected from Wikipedia snapshots dated February 2019.
10. **ARCD[13]:** The Arabic Reading Comprehension Dataset (ARCD) introduced in 2019 and consists of two main components. The first is a collection of 1,395 questions crafted by crowdsourced workers based on Wikipedia articles. The second component is an Arabic translation of the Stanford Question Answering Dataset, commonly referred to as Arabic-SQuAD.

[5] https://aclanthology.org/W19-4619/
[6] https://catalog.ldc.upenn.edu/LDC2003T12
[7] https://oscar-project.org/
[8] https://archive.org/details/arwiki-20190201
[9] https://data.baai.ac.cn/details/ArabicText-2022
[10] https://www.tensorflow.org/datasets/catalog/c4
[11] https://www.kaggle.com/datasets/mahmoudqaddoumi/arabic-library
[12] https://github.com/google-research-datasets/tydiqa?tab=readme-ov-file#download-the-dataset
[13] https://metatext.io/datasets/arabic-reading-comprehension-dataset-(arcd)

11. **CC100-Arabic[14]:** It was introduced in 2020 and is part of a larger collection of 100 monolingual corpora. This Arabic dataset was extracted and processed from the Common Crawl snapshots spanning January to December 2018, utilizing the CC-Net repository. The corpus, presented in text file format, comprises 5.4 BB of Arabic data.
12. **OpenSubtitles2016 corpus[15]:** It is an extensive collection of parallel corpora extracted from movie and TV subtitles, containing 2.6 billion sentences across 60 languages. This dataset contains 1689 language pairs of text and 3.36 million subtitle files, covering over 152,939 unique movies and TV episodes sourced from OpenSubtitles.org. The Arabic section of this corpus grew as the number of subtitle files increased from 42,400 in 2012+2013 release to 67,300 in 2016. This Arabic subset covered 34,100 unique IMDb titles and contains 60.0 million sentences, totaling 327 million tokens.
13. **AraNews[16]:** It is a large-scale Arabic news dataset crawled during late 2019 and early 2020. It contains 5,187,957 news articles collected from 50 newspapers across 15 Arab countries, the USA, and the UK. The dataset covers a wide range of topics, with articles categorized into 17 thematic categories, including Politics, History, Society, Media, Entertainment, Weather, Sports, Social Media, Heath, Culture and Art, Economy, Religion, Education, Technology, Fashion, Local News, and International News.
14. **Hindawi[17]:** This dataset offers a collection of literary works written in MSA form. It includes metadata for each book, such as the title, author, abstract, and link to access the complete text online. The dataset contains over 3,000 books, totaling more than 120 million tokens, with a size of approximately 476 MB. However, the exact date range for the collected data is not explicitly documented.
15. **CulturaX[18]:** It is a multilingual dataset designed for training LLMs across 167 languages. It contains 6.3 trillion tokens and was constructed by merging mC4 (version 3.1.0) and all available OSCAR corpora from different years (OSCAR 2019, 2020, 2021, and 2022). The Arabic subset contains around 74K documents, which results in more than 69.35 tokens.
16. **SANAD[19]:** The Single-label Arabic News Articles Dataset (SANAD) is a large-scale Arabic news corpus designed for automatic text categorization. The dataset contains 194,797 news articles collected from three major Arabic news portals: AlKhaleej, AlArabiya, and Akhbarona-Alanba. The articles are categorized into seven distinct categories: Finance, Sports, Culture, Technology, Politics, Medical, and Religion. Each article is assigned a single label based on its original news source categorization. The dataset spans a period from 2008 to 2018, with data collected from AlKhaleej (2008–2018), AlArabiya (October 2012 – April 2018), and Akhbarona-Alanba (January 2011 – October 2018), covering a decade of Arabic news articles.
17. **Arabic E-book corpus[20]:** It is a freely accessible dataset of 1,745 books released by the Hindawi Foundation between 2008 and 2024, with around 280 million tokens. The books cover wide range of categories, such as plays, poetry, children's books, novels, and non-fiction.

---

[14] https://metatext.io/datasets/cc100-arabic
[15] https://www.opensubtitles.org/en/search/subs
[16] https://github.com/UBC-NLP/wanlp2020_arabic_fake_news_detection
[17] https://huggingface.co/datasets/alielfilali01/Hindawi-Books-dataset
[18] https://huggingface.co/datasets/uonlp/CulturaX
[19] https://data.mendeley.com/datasets/57zpx667y9/2
[20] https://huggingface.co/datasets/mohres/The_Arabic_E-Book_Corpus

### 4.4 Dialectal Arabic (DA) Resources

Dialectal Arabic (DA) refers to the diverse spoken forms of Arabic that are used in everyday communication throughout the Arab world. These dialects differ significantly from MSA across several linguistic dimensions, including phonology, morphology, orthography, and syntax [35]. DA is typically categorized by region into major groups, such as Egyptian, North African, Levantine, Gulf, and Yemeni dialects. Each of these groups contains further sub-varieties, including specific dialects such as Tunisian, Algerian, Lebanese, Syrian, Jordanian, Saudi, and Qatari [36].

Given the widespread use of DA in daily conversations, integrating DA resources into the training of LLMs is essential for developing models capable of understanding and generating Arabic, as it is spoken in real-life contexts. Table 5 provides an overview of the models pre-trained on DA resources.

Table 5: DA Datasets for Arabic LLMs. Data as of February 2025.

| Model Name | Dialect | Dataset Size | Tokens/ Vocabulary size | Data Source |
|---|---|---|---|---|
| CAMeLBERT-DA [19] | Egyptian, Shami, Gulf Arabic, Iraqi, Saudi, Algerian, Tunisian, Morocco, Lebanese, Sudanese, Libyan, and Yemeni | 54GB | 5.8 billion words | CALLHOME Egyptian Arabic Transcripts, CALLHOME Egyptian Arabic Transcripts Supplement, Babylon Levantine Arabic Transcripts, Levantine Arabic QT Training Data Set 4 Transcripts, Levantine Arabic QT Training Data Set 5 Transcripts, Gulf Arabic Conversational Telephone Transcripts, Iraqi Arabic Conversational Telephone Transcripts, Levantine Arabic Conversational Telephone Transcripts, Fisher Levantine Arabic Conversational Telephone Transcripts, AOC dataset, Arabic-Dialect/English Parallel Text, Arabic Multi Dialect Text Corpora, A Multidialectal Parallel Corpus of Arabic, Multi-Dialect - Multi-Genre Corpus of Informal Written Arabic, Youtube Dialectal Arabic Comment Corpus (YouDACC), PADIC Corpus, Curras Corpus, WERd, BOLT Egyptian Arabic SMS/Chat and Transliteration, SDC, SUAR Corpus, Arap-Tweet, Gumar, MADAR Corpus, Habibi Corpus, NADI 2020 Corpus, and QADI Corpus |
| SudaBERT [20] | Sudanese | ~ 13 million sentences | Not explicitly mentioned | Twitter and public Telegram channels |
| AraRoBERTa-SA [16] | Saudi | 2.42 million tweets | 45.4 million tokens | Twitter |
| AraRoBERTa-EG [16] | Egyptian | 2.10 million tweets | 37.2 million tokens | Twitter |
| AraRoBERTa-KU [16] | Kuwaiti | 478K tweets | 8.9 million tokens | Twitter |
| AraRoBERTa-OM [20] | Omani | 200K tweets | 3.8 million tokens | Twitter |
| AraRoBERTa-LB [16] | Lebanese | 204K tweets | 3.6 million tokens | Twitter |
| AraRoBERTa-JO [16] | Jordanian | 97K tweets | 2.6 million tokens | Twitter |
| AraRoBERTa-DZ [16] | Algerian | 103K tweets | 1.9 million tokens | Twitter |

| Model Name | Dialect | Dataset Size | Tokens/ Vocabulary size | Data Source |
|---|---|---|---|---|
| DziriBERT [23] | Algerian | 150 MB | 20 million tokens | Twitter |
| TunBERT [25] | Tunisian | 67.2 MB | 8.2K tokens | Tunisian Common-Crawl-based dataset |
| MorRoBERTa [26] | Moroccan | ~497.61 MB | ~71 billion tokens | Facebook, YouTube, Twitter |
| MorrBERT [26] | Moroccan | ~497.61 MB | ~71 billion tokens | Facebook, YouTube, Twitter |
| SaudiBERT [27] | Saudi | 26.3 GB | Not explicitly mentioned | SFC and STMC Corpus |
| AlcLaM [28] | Not explicitly mentioned | ~ 3.37 million dialectal sentences | ~ 54.56 million tokens | comments on Arabic pages on Facebook and videos from Arabic-speaking YouTubers. |
| EgyBERT [29] | Egyptian | 10.4 GB | 75K vocabulary | EFC and Egyptian Tweets Corpus |
| DarijaBERT [31] | Moroccan (Darija) | 691 MB | 80K tokens | YouTube, 9essas, and Twitter |
| DarijaBERT-arabizi [31] | Moroccan (Darija) Latin script (Arabizi) | 287 MB | 110K tokens | Arabizi Dataset |
| DarijaBERT-mix [31] | Moroccan Arabic (Darija Arabizi) | 14 million sequences (9.5 million sequences in Darija dialect and 4.6 million in Arabizi) | 160k tokens | Combined dataset from DarijaBERT and DarijaBERT-arabizi |
| Atlas-Chat [43] | Moroccan (Darija) | 458K instruction samples | Not explicitly mentioned | Darija-SFT-Mixture dataset |

The available datasets used to pre-train the DA language models are as follows:

1. **CALLHOME Egyptian Arabic Transcripts[21]:** This corpus contains transcripts from 120 telephone conversations between native Egyptian Arabic speakers. Each transcript represents a continuous segment of either five or ten minutes from a conversation.
2. **CALLHOME Egyptian Arabic Transcripts Supplement[22]:** This additional corpus, developed by the LDC, expands the original CALLHOME collection by providing transcripts for 20 more telephone conversations in Egyptian Arabic.
3. **Babylon Levantine Arabic Transcripts[23]:** A collaborative effort by BBN Technologies and the American University of Beirut, this corpus includes 60.6 hours of natural Shami dialects speech (Jordanian, Lebanese,

[21] https://catalog.ldc.upenn.edu/LDC97T19
[22] https://catalog.ldc.upenn.edu/LDC2002T38
[23] https://catalog.ldc.upenn.edu/LDC2005S08

Palestinian, and Syrian). It contains 76,227 statements, each accompanied by a corresponding audio recording and a written transcript.

4. **Levantine Arabic QT Training Data Set 4 Transcripts[24]:** Developed by the LDC, it contains approximately 138 hours of conversational telephone speech in Shami dialects along with corresponding transcripts. It comprises 901 separate phone calls, predominantly of Lebanese speakers, as well as other Shami dialects. The dataset encompasses 1,802 calls from 665 female and 1,137 male speakers.
5. **Levantine Arabic QT Training Data Set 5 Transcripts[25]:** Also developed by the LDC, this corpus offers transcripts for about 250 hours of telephone conversations in Shami dialects. The dataset comprises 1,660 calls recorded between 2003 and 2005. While Lebanese speakers form the majority, they also include the Jordanian, Palestinian, and Syrian dialects.
6. **Gulf Arabic Conversational Telephone Transcripts[26]:** Created by Appen Pty Ltd. in Sydney, Australia, in 2004, this corpus contains transcripts of approximately 2,800 minutes of natural telephone conversations in Gulf dialects. It includes 976 conversational sides from 975 distinct Gulf Arabic speakers.
7. **Iraqi Arabic Conversational Telephone Transcripts[27]:** Developed by Appen Pty Ltd. between 2003 and 2004, this corpus contains transcripts for approximately 3,000 minutes of speech in the Iraqi dialect. It includes 478 conversations from 474 unique speakers, with an average call duration of approximately six minutes.
8. **Levantine Arabic Conversational Telephone Transcripts[28]:** This dataset contains random phone dialogues in the Shami dialects, including 982 unique speakers and 985 conversations. Each conversation typically lasted between five and six minutes.
9. **Fisher Levantine Arabic Conversational Telephone Transcripts[29]:** This corpus consists of transcripts from 279 telephone conversations in Shami dialects. Most speakers were from Jordan, Lebanon, and Palestine.
10. **AOC Dataset[30]:** The Arabic Online Commentary Dataset (AOC) was collected from April to early October 2010 by crawling the websites of three Arabic newspapers and extracting online articles and their associated reader comments. The dataset includes content from:
    a. Al-Ghad (الغد), a Jordanian newspaper.
    b. Al-Riyadh (الرياض), a Saudi newspaper.
    c. Al-Youm Al-Sabe' (اليوم السابع), an Egyptian newspaper.
11. **Arabic-Dialect/English Parallel Text[31]:** Developed by Raytheon BBN Technologies, LDC, and Sakhr Software, this dataset contains approximately 3.5 million tokens of Shami and Egyptian dialect sentences paired with their English translations.
12. **PADIC Corpus[32]:** The Parallel Arabic Dialect Corpus (PADIC) is a multi-dialectal dataset created as part of the "TORJMAN" National Research Project. It was supervised by the Scientific and Technical Research Center for the Development of Arabic Language and supported by Algeria's Ministry of Higher Education and Scientific

---

[24] https://catalog.ldc.upenn.edu/LDC2005S14
[25] https://catalog.ldc.upenn.edu/LDC2006T07
[26] https://catalog.ldc.upenn.edu/LDC2006T15
[27] https://catalog.ldc.upenn.edu/LDC2006T16
[28] https://catalog.ldc.upenn.edu/LDC2007T01
[29] https://catalog.ldc.upenn.edu/LDC2007T04
[30] https://www.cs.jhu.edu/data-archive/AOC-2010/?C=D;O=D
[31] https://catalog.ldc.upenn.edu/LDC2012T09
[32] https://sourceforge.net/projects/padic/

Research. The PADIC corpus contains over 6,000 sentences from six dialects: two from Algeria (Algiers and Annaba), Palestinian, Syrian, Tunisian, Moroccan, and MSA.

13. **Curras Corpus[33]:** This Palestinian dialect corpus contains approximately 56,000 tokens. The dataset was collected from a diverse range of sources including Facebook, Twitter, Palestinian stories, TV shows, and other textual resources.
14. **BOLT Egyptian Arabic SMS/Chat and Transliteration[34]:** Collected by the LDC, this dataset captures authentic Short Message Service (SMS) and chat communications in the Egyptian dialect. It includes 5,691 conversations, totaling 1,029,248 words across 262,026 messages.
15. **SDC[35]:** The Shami Dialect Corpus (SDC) is the first collection of Levantine Arabic dialects (Shami dialects), comprising 117,805 sentences and 13,151,489 tokens. It was gathered using both automated methods (Twitter API) and manual collections from online blogs and forums.
16. **Gumar Corpus[36]:** This is a morphologically annotated collection of Gulf Arabic texts containing over 112 million words from more than 1,200 documents. The resource primarily includes long anonymously published online conversational novels, often with romantic themes, drama, and tragedy, providing a rich mix of dialectal Gulf Arabic and MSA.
17. **MADAR Corpus[37]:** The Multi-Arabic Dialect Applications and Resources (MADAR) corpus is a parallel dataset of sentences translated from the Basic Traveling Expression Corpus (BTEC) into 25 Arabic city dialects, along with MSA, English, and French versions. It consists of CORPUS-25 (2,000 sentences in 25 dialects) and CORPUS-5 (10,000 sentences in five key dialects: Doha, Beirut, Cairo, Tunis, and Rabat), totaling 100,000 dialectal Arabic sentences. The 25 cities covered included Morocco (Rabat and Fes), Algiers, Tunisia (Tunis and Sfax), Libya (Tripoli and Benghazi), Egypt (Cairo, Alexandria, and Aswan), Khartoum, Jerusalem, Jordan (Amman and Salt), Beirut, Syria (Damascus and Aleppo), Iraq (Mosul, Baghdad, and Basra), Doha, Muscat, Saudi Arabia (Riyadh and Jeddah), and Sana'a.
18. **Habibi Corpus[38]:** This multi-dialect, multi-national Arabic song lyrics corpus contains over 30,000 songs from 18 different Arab countries. It includes 30,072 song titles, 527,870 sentences, and more than 3.5 million words, covering a wide range of artists from various regions, categorized into six main dialects: Egyptian, Gulf, Levantine, Iraqi, Sudanese, and Maghrebi (North African). The 18 countries covered include Egypt, Saudi Arabia, Lebanon, Iraq, Sudan, Kuwait, Syria, UAE, Morocco, Tunisia, Yemen, Jordan, Algeria, Qatar, Bahrain, Oman, Palestine, and Libya. These countries are categorized into six main dialects: Egyptian, Gulf, Shami, Iraqi, Sudanese, and Maghrebi.
19. **NADI 2020 Corpus[39]:** The Nuanced Arabic Dialect Identification (NADI) corpus was developed for the NADI Shared Task, consisting of 30,957 tweets collected from Twitter over 10 months in 2019. It represents 21 Arab countries and 100 provinces and supports two subtasks: country-level dialect identification and province-level dialect identification. The 21 countries covered include Algeria, Bahrain, Djibouti, Egypt, Iraq, Jordan, Kuwait,

33 https://www.jarrar.info/publications/JHRAZ17.pdf
34 https://catalog.ldc.upenn.edu/LDC2017T07
35 https://github.com/GU-CLASP/shami-corpus
36 https://camel.abudhabi.nyu.edu/gumar/
37 https://sites.google.com/nyu.edu/madar/home#h.xpcfdhjyc95c
38 http://ucrel-web.lancaster.ac.uk/habibi/
39 https://sites.google.com/view/nadi-shared-task

Lebanon, Libya, Mauritania, Morocco, Oman, Palestine, Qatar, Saudi Arabia, Somalia, Sudan, Syria, Tunisia, UAE, and Yemen.

20. **QADI Corpus[40]:** The QCRI Arabic Dialects Identification (QADI) corpus contains over 540,000 dialectal tweets from 2,525 Twitter accounts across 18 Arab countries. Collected via the Twitter API during March and April 2018, it focuses on dialectal content while excluding users who primarily tweeted in MSA. The dataset ensures a balanced representation across countries, with an average of 140 accounts and 30,000 tweets per country. The 18 countries covered include Algeria, Bahrain, Egypt, Iraq, Jordan, Kuwait, Lebanon, Libya, Morocco, Oman, Palestine, Qatar, Saudi Arabia, Sudan, Syria, Tunisia, UAE, and Yemen.
21. **Darija-SFT-Mixture dataset[41]:** The Darija-SFT-Mixture dataset contains 458,000 instruction samples. It integrates various resources, including existing Darija language datasets such as the Darija Open Dataset (DODa-10K), MADAR, FLORES+, MArSum, MW-QA, MSM-MG, and NLLB-Seed, along with novel synthetic datasets and translations of English instructions. This diverse mixture was designed to enhance the ability of the model to understand and generate Moroccan Arabic (Darija).

### 4.5 Combination of Arabic Forms Resources

Although many Arabic LLMs focus on specific forms of Arabic, such as MSA or DA, some models leverage a combination of these forms to create more adaptable and realistic language models. This approach mirrors real-world language use, where Arabic speakers often transition between different forms, depending on the context. For instance, in social media conversations, users may alternate between MSA and DA, while CA, although less frequently used, might appear in religious or classical citations. In addition, in multicultural settings, speakers sometimes switch between different dialects, reflecting the diverse linguistic landscape of the Arab world. By combining multiple forms of Arabic, these models aim to handle a broader range of tasks and more accurately represent the complex interplay between linguistic forms in daily life. This approach aligns with the natural way in which Arabic is used in real-life scenarios. Table 6 provides an overview of Arabic language models that have been pre-trained using various combinations of Arabic forms, including MSA, DA, and CA. These models include a wide range of data sources, such as social media, news articles, literary texts, and web-crawled data, to enhance their adaptability and relevance in real-world applications. By leveraging multiple Arabic forms, these models offer broader linguistic coverage, capturing the dynamic nature of Arabic language use in different contexts.

Table 6: Combinations of Multiple Arabic Forms for Arabic LLMs. Data as of February 2025.

| Model Name | Forms | Dataset Size | Tokens/Number of words | Data Source |
|---|---|---|---|---|
| QARiB [3] | MSA and DA | 120M sentences and tweets | 2.7B Arabic words | Arabic GigaWord corpus, Abulkhair Arabic Corpus, OpenSubtitles 2016, and tweets |
| ArabicBERT [18] | MSA and DA | Not explicitly mentioned | 8.2 billion words | OSCAR corpus (unshuffled) and Wikipedia dump for Arabic |
| CAMeLBERT-Mix [19] | CA, MSA, and DA | 167GB | 17.3 billion words | All datasets that were used in CAMeLBERT-MSA, CAMeLBERT-DA, and CAMeLBERT-CA |

[40] https://alt.qcri.org/resources/qadi/
[41] https://huggingface.co/datasets/MBZUAI-Paris/Darija-SFT-Mixture

| | | | | |
|---|---|---|---|---|
| ARBERT [4] | MSA and Egyptian DA | 61GB | 6.5 billion tokens | Books (Hindawi), El-Khair, Gigawords, OSIAN, OSCAR-MSA, OSCAR-Egyptian, and Wiki |
| MARBERT [4] | MSA and DA | 61GB | 15.6 billion tokens | 1B Arabic tweets sampled from a larger dataset of 6B tweets |
| AraT5 [47] | MSA, Saudi, Egyptian, Bahraini, Kuwaiti, Qatari, Emirati, Omani, Yemeni, Libyan, Tunisian, Iraqi, Sudanese, Shami, Algerian, and Moroccan | 248 GB (70GB for MSA and 178GB for DA) | 29 billion tokens | AraNews, Books, El-Khair, Gigawords, OSIAN, OSCAR-MSA, OSCAR-Egyptian, Twitter, and Wiki |
| AraMUS [46] | MSA and DA | 529GB | Not explicitly mentioned | Common Crawl, OSIAN, El-KHAIR corpora, Arabic dialect textual data, and others |
| JASMINE [35] | MSA, CA, and DA | 421.8GB (222.8GB for MSA, 12GB for CA, and 178GB for DA) | 46.7 billion tokens (23.7B for MSA, 1.1B for CA, and 21.9B for DA) | OpenITI v1.6, AraNews, El-Khair, Gigaword, OSCAR, OSIAN, Wikipedia Arabic, Hindawi Books, AraC4, and Twitter |
| AraPoemBERT [30] | CA and MSA | 182 MB | 29 million tokens | Arabic poetry text collected from online sources (Almausua and Aldiwan) |
| AraStories [50] | MSA, Egyptian and Moroccan | 548K stories | Not explicitly mentioned | Stories generated by GPT-4 and translated stories from the TinyStories dataset |
| Fanar [41] | MSA and CA | Not explicitly mentioned | ~ 1 Tira tokens (410.3 billion for Arabic, 512.9 for English, and 102.6 for GitHub code) | Web documents, scientific articles, encyclopedic entries, mathematical problems, books, news articles, and source code from common programming languages. |

### 4.6 Discussion

The datasets used for pre-training Arabic LLMs reflect the rich diversity of the Arabic language and its dialects. However, challenges remain, particularly in representing DA and CA forms. As shown in Figure 2, existing LLMs primarily focus on MSA, which forms the foundation of many formal texts across the Arab world, including news articles, academic works, and official documents. Datasets such as the 1.5 billion Words Corpus, the OSIAN Corpus, and the Arabic Wikipedia dump form the core resources for MSA-based LLMs, as well as for models that incorporate combinations of MSA with other forms of Arabic. Figure 3 illustrates the frequency with which different datasets are used across models, showing the critical role of these resources in LLM development.

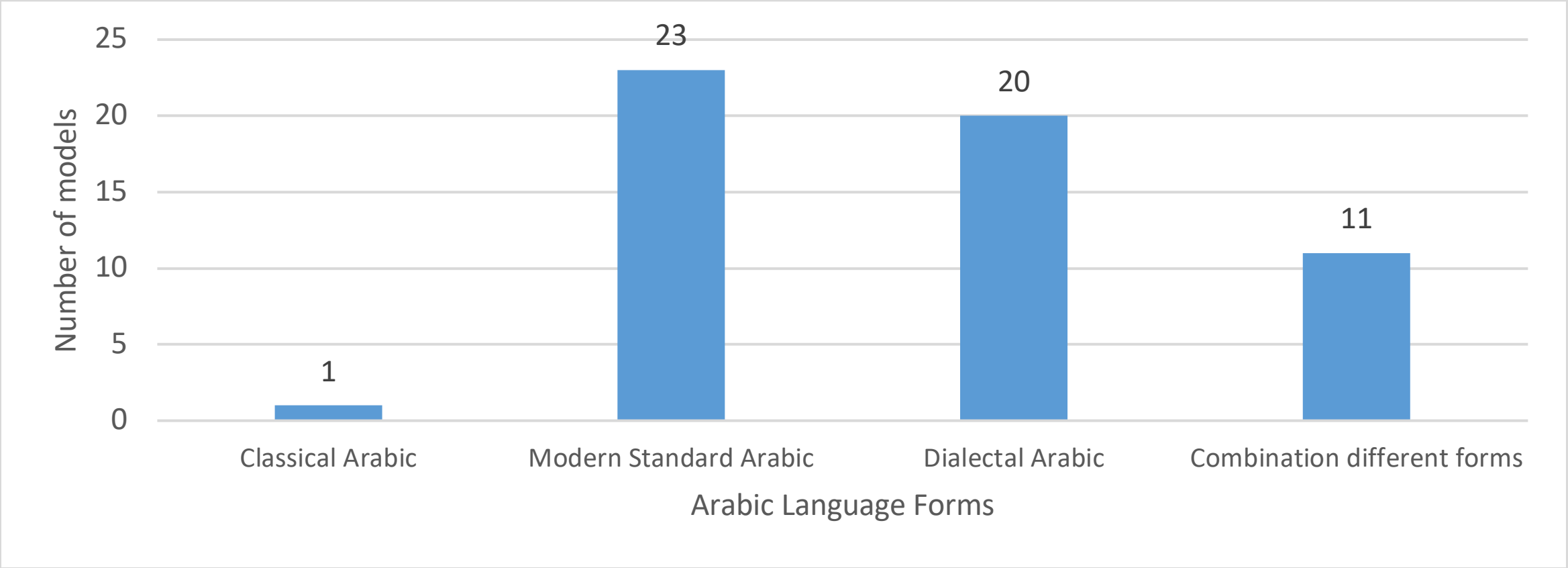

Figure 2: Distribution of Models Across Different Arabic Language Forms. Data as of February 2025

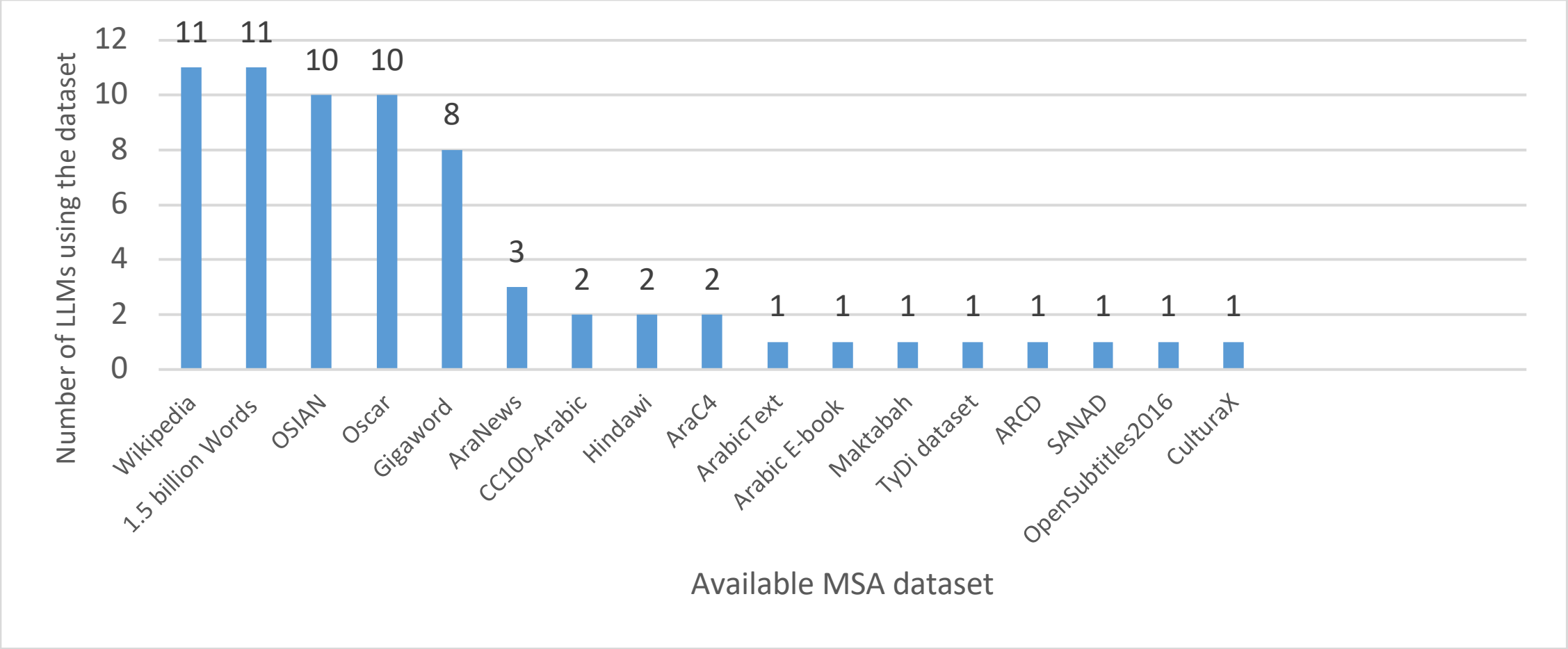

Figure 3: Frequency of Dataset Usage in LLMs Including MSA. Data as of February 2025.

On the other hand, DA, commonly used in everyday conversation and social media, presents a unique challenge owing to its wide regional diversity and limited scope in terms of available datasets. Building more dialect-inclusive models is essential for addressing this variation. The primary sources for DA datasets are social media platforms, such as X (previously Twitter), YouTube, and Facebook, which are vital for capturing informal, everyday spoken language. Figure 4 highlights the distribution of major Arabic dialects within the dialectal Arabic language models. Notably, North African dialects are represented by 10 models, with eight of those specifically focusing on Moroccan Arabic. In contrast, Iraqi and Yemeni Arabic dialects are among the least represented, with only two models each. However, compared with DA, CA remains a greater challenge owing to its limited digital resources. Only a few models, including CAMeLBERT-CA, JASMINE, Fanar, and AraPoemBERT, have focused on CA, reflecting the need for more effort in this area.

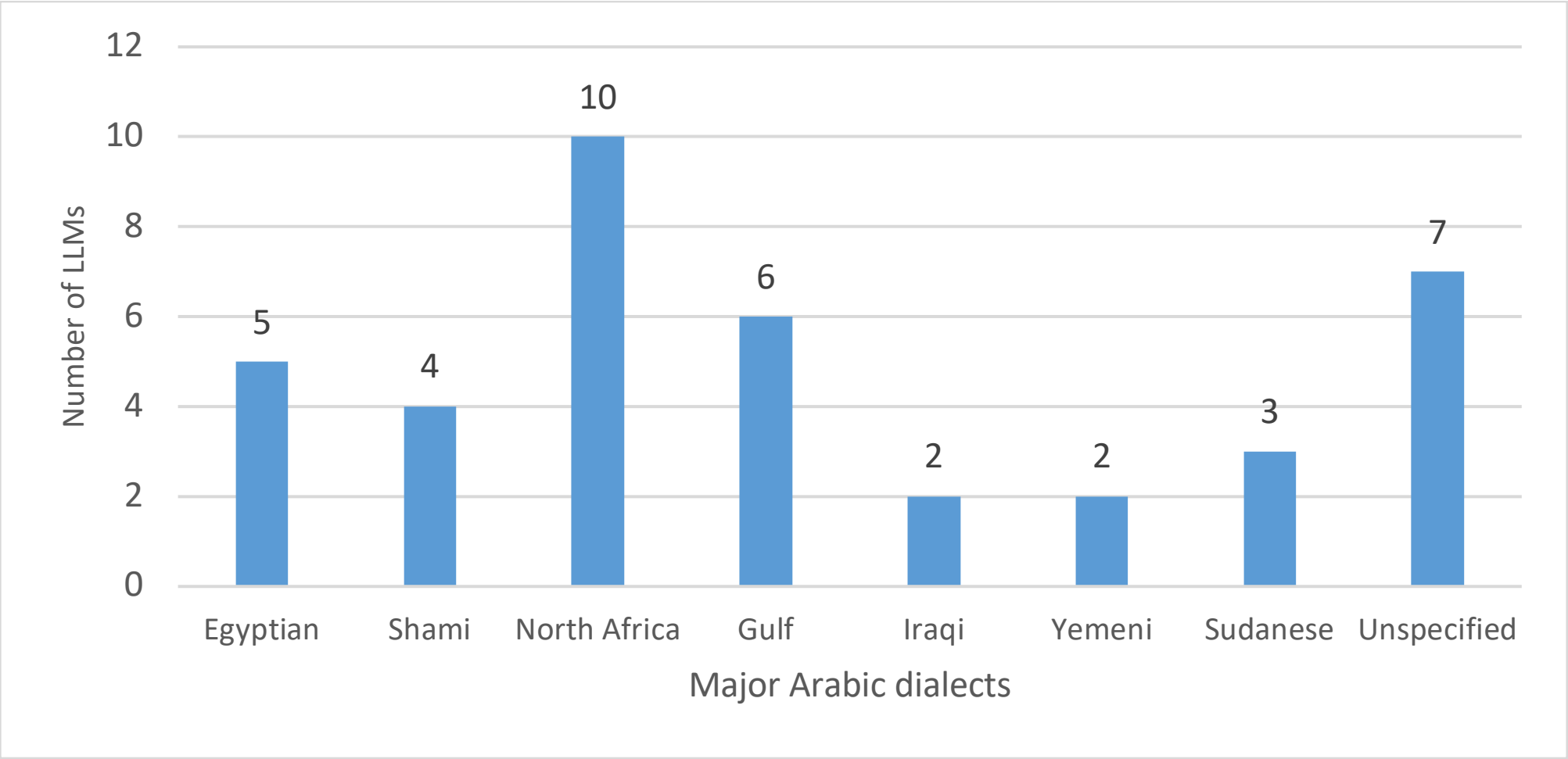

Figure 4: Distribution of LLMs Across Major Arabic Dialects. Data as of February 2025.

Finally, the models trained on a combination of Arabic forms (CA, MSA, and DA) highlight the potential for more adaptable language models. These models, such as CAMeLBERT-Mix and JASMINE, represent the natural linguistic code-switching that occurs in Arabic-speaking communities, where speakers often shift between different forms, depending on the context. The integration of multiple Arabic forms in pre-training broadens the applicability of these models, making them more adaptable for tasks requiring both formal and informal language understanding.

## 5 ARABIC LLMS

This section provides an overview of monolingual, bilingual, and multilingual LLMs for the Arabic language and its dialects.

### 5.1 Monolingual Models

The development of monolingual Arabic language models has seen significant progress since 2020, with models based on various architectures, such as BERT, GPT-2, ELECTRA, T5, and RoBERTa. These models were trained exclusively on Arabic text, allowing them to learn the nuances of a language's morphology, syntax, semantics, and dialectical variations. This section provides a comprehensive overview of the 26 monolingual Arabic models, discussing their architecture, the variety of downstream tasks they are designed to handle, and the baseline models used for performance evaluation.

We categorized these models into three architectural groups: BERT-based models, with a specific focus on Arabic dialectal BERT-based models, GPT-based decoders, and other architectures such as ELECTRA and T5. For each model, we highlight key details, including parameter size, training data size, variants based on size or data, intended applications, and comparative evaluation results on tasks such as sentiment analysis, named entity recognition, question answering, and dialect identification. Quantitative metrics, such as accuracy, F1 scores, and perplexity, provide tangible benchmarks for the models. We also discuss the limitations and weaknesses uncovered through the model evaluations to provide a balanced perspective.

#### *5.1.1BERT-Based Models*

AraBERT [2], introduced in 2020, was the first known BERT-based model pre-trained for the Arabic language. The model followed the same BERT architecture with approximately 110M parameters and was trained on 70M Arabic sentences. The model has two variants: AraBERT-v.01, which does not require sub-word unit segmentation during training, and AraBERT-v1, which was trained using sub-word segmentation. Both variants were fine-tuned for Sequence Classification, Named Entity Recognition (NER), and QA, and were evaluated on Sentiment Analysis (SA), NER, and QA tasks and compared to the multilingual BERT, namely mBERT [5]. The results showed that both versions of AraBERT outperformed mBERT on SA and NER tasks, particularly in dialects, noting that AraBERT was trained on MSA data only. For the QA task, AraBERT had a better F1 score and sentence match, but a lower score on the exact match metric.

The success of AraBERT has paved the way for subsequent BERT-based models tailored for Arabic. MARBERT and ARBERT [4], released later in 2020, share the same BERT-based architecture but differ in their training data. ARBERT focused on MSA, while MARBERT incorporated both MSA and dialectal Arabic data. For the evaluation, the authors compared both models to the multilingual models (mBERT and XLM [56]) and AraBERT across five downstream tasks. The results revealed MARBERT's superiority in SA, Social Meaning, and Dialect Identification, whereas ARBERT excelled in Topic Classification. Both models performed equally well in the NER tasks. Overall, the evaluation demonstrated that ARBERT excelled in formal Arabic tasks, whereas MARBERT showed superior performance in dialectal Arabic and social media-related tasks.

To address the diversity of the Arabic language, the CAMeLBERT [19] family of models was introduced in 2021. This suite includes variants trained on MSA, DA, CA, and a combination of these. This approach allowed for a more nuanced handling of different Arabic varieties in tasks such as NER, Part-of-Speech (POS) tagging, and dialect identification. The fine-tuning results showed that CAMeLBERT-MSA performed best in NER, POS Tagging, and SA tasks, while CAMeLBERT-DA excelled in dialect identification, and CAMeLBERT-CA outperformed other models in poetry classification. Notably, the larger dataset size of CAMeLBERT-MSA may have contributed to its strong performance across tasks. Eight baseline models were used to compare the CAMeLBERT variants: mBERT, AraBERT, ArabicBERT, multidialect-Arabic-BERT [58], GigaBERTv4, MARBERT, and ARBERT. Comparing CAMeLBERT to other baseline models, the best model for most tasks was AraBERT, followed by CAMeLBERT-mix.

JABER and SABER [22] were released as Junior (12-layer) and Senior (24-layer) Arabic BERT-based and BERT-large language models, respectively. The models were trained on 115GB of Arabic corpus and evaluated on the following downstream tasks: 4 single-sentence, 2 sentence-pair, one multi-label classification, and single regression tasks. The authors compared their model to the following baseline models: Arabic-BERT, AraBERT, CAMeLBERT, ARBERT, and MARBERT. The results showed that SABER outperformed previous models across all tasks, whereas JABER showed strength in single and sentence-pair classification tasks.

Several domain-specific BERT models have been developed. QARiB [3], trained on a mixture of MSA and dialectal Arabic, focused on text categorization. AraLegal-BERT, with base (12-layer) and large (24-layer) variants, specialized in Arabic legal texts. Trained on 13.7 million sentences of judicial, legal, and Islamic documents, it outperformed AraBERT, ARBERT, and mBERT in legal text classification, keyword extraction, and NER tasks. Similarly, AraPoemBERT focused on Arabic poetry, excelling in tasks such as SA, poetry meter classification, and poet's gender classification.

The evolution of BERT-based models for Arabic has been marked by significant progress, moving from the initial focus on MSA to specialized versions targeting classical and dialectal Arabic as well as domain-specific applications such as legal texts and poetry. A notable trend is the positive correlation between model complexity and performance, as exemplified by models such as SABER, which leverage larger datasets and more complex architectures to achieve superior results across various classification tasks. Although different models excel in different areas, SA and NER have emerged as popular tasks for fine-tuning and benchmarking. The rapid progress of the field is evident in the consistent pattern of new models outperforming their predecessors across most tasks, demonstrating the effectiveness of iterative refinement in model design.

#### *5.1.2Dialectical BERT-Based Models*

Linguistic diversity across the Arab world has prompted the development of dialect-specific BERT-based models to address the unique characteristics of various Arabic dialects. For instance, SudaBERT [20] focused on Sudanese Arabic, utilizing the Arabic-BERT architecture, and trained on 13 million sentences in the Sudanese dialect. The model was compared with Arabic-BERT on SA and NER tasks, and the results showed that SudaBERT outperformed Arabic-BERT in the SA task in the Sudanese dialect, whereas Arabic-BERT performed better in SA and NER in MSA.

AraRoBERTa [16] is a dialect-specific language model trained using the RoBERTa-base configuration for seven dialects, Saudi, Egyptian, Kuwaiti, Omani, Lebanese, Joradanian, and Algerian, where each dialect is used to train a separate model. The baseline models used to compare the performance of the models were mBERT, XLM-R, and AraBERT. The models were trained for dialect classification using fully, semi-supervised, and weakly supervised approaches. For the fully supervised approach, in addition to the baselines, the model was compared to a traditional logistic regression model, which yielded the best results in the Kuwaiti, Lebanese, Jordanian, and Algerian dialects, whereas AraRoBERTa performed best in the Saudi and Egyptian dialects because of their larger data size compared to other dialects. The semi-supervised approach obtained better results than the supervised approach in Egyptian, Omani, Lebanese, and Algerian dialects. However, the weakly supervised approach performed poorly compared with the other approaches.

The DziriBERT [23] is specialized in Algerian Arabic, trained on a dataset of 1.1 million tweets in the Algerian dialect and has the same architecture as the BERT-base model. The authors have compared DziriBERT to multilingual transformers (mBERT and XLM-R) and multiple standard and dialectal Arabic models (AraBERT, QARiB, CAMeLBERT-DA, CAMeLBERT-mix, and MARBERT). The results show that DziriBERT obtained the best results in SA, Emotion Classification, and Topic Classification tasks. The MARBERT model was the second-best in SA and Emotion Classification.

TunBERT [25] targets Tunisian Arabic using a corpus of 500k sentences extracted from Tunisian social media, blogs, and websites. The authors implemented TunBERT with two approaches: the first approach, namely TunBERT-P, relies on BERT Pytorch implementation, and the second approach, TunBERT-T, relies on Tensorflow implementation. The authors have compared TunBERT to mBERT, AraBERT, GigaBERT, and MARBERT on SA, Dialect Identification, and Reading Comprehension QA tasks. The results show that TunBERT-P achieved the best performance in SA and dialect identification but failed to perform better than AraBERT and GigaBERT in the Reading Comprehension QA task. In addition, TunBERT-P performed better than TunBERT-T for all three tasks.

The Moroccan Arabic dialect (Darija) is popular and has attracted the attention of several researchers. Several recent language models have focused on Darija, including MorrBERT [26], DarijaBERT [31], and Atlas-Chat [43]. MorrBERT is a BERT-based model trained on over 71 billion tokens of social media data consisting of 125 million parameters. It was evaluated on sentiment analysis, dialect identification, and language classification, and compared well with models such as mBERT and XLM-R. However, MARBERT performed the best in sentiment analysis. The authors of MorrBERT also released MorRoBERTa, a RoBERTa-based model with around 83.5M parameters and was trained on the same data as MorrBERT. MorRoBERTa achieved higher accuracy results than MorrBERT and the other baseline models in SA and dialect identification tasks.

DarijaBERT has three variants covering Arabic letters (DarijaBERT), Latin letters (DarijaBERT-arabizi), and both (DarijaBERT-mix). The three models utilized a BERT-based architecture. The models were evaluated on four downstream tasks: Dialect Identification, SA, Sarcasm Detection, and Topic Classification, and they were compared to XLM-RoBERTa-Base, mBERT, AraBERTv0.2, CAMeLBERT-DA, QARiB, and MARBERT. For the Dialect Identification task, DarijaBERT-mix outperformed all models including DarijaBERT, followed by DarijaBERT then MARBERT. For the SA task, lower scores were obtained compared to the Dialect Identification task, with DarijaBERT outperforming the other models. However, for the Sarcasm Detection task, MARBERT obtained the best F1 score. For the Topic Classification task, DarijaBERT-mix obtained the best results, followed by DarijaBERT. The authors concluded that models trained on the Arabic dialect performed better than those trained on MSA.

Atlas-Chat is the latest known LLM, designed specifically for Moroccan Arabic (Darija). Atlas-Chat was created using instruction-tuning on Gemma-2 base model, with two size variants: Atlas-Chat-2B and Atlas-Chat-9B, built upon Gemma-2-2B and Gemma-2-9B, respectively. The model was trained on datasets targeting various natural language processing tasks, including translation, sentiment analysis, question answering, summarization, and story completion. Interestingly, the authors experimented with fine-tuning both an instruction-tuned model and a base model, and their results indicated that continual fine-tuning of the instruction-tuned Gemma-2 model yielded higher performance scores compared to other settings. In a comparative evaluation, Atlas-Chat was benchmarked against multiple Arabic and multilingual base models such as Jais, AceGPT, and Llama. The comparison results revealed that the larger Atlas-Chat-9B variant excelled in sentiment analysis and translation tasks, whereas Llama demonstrated superior performance in text summarization.

AlcLAM [28] was developed for Arabic dialects, with emphasis on dialect identification and offensive language detection. AlcLAM has the same architecture as BERT and is trained on more than 3 million sentences of Arabic dialects. The authors compared AlcLAM to mBERT, LaBSE [60], AraBERT, ArBERT, Md-BERT [58], MARBERT, and CAMeL on SA, Offensive Language Detection, Dialect Identification. The results show that AlcLAM performed best in the majority of dialect identification and SA tasks and excelled in Offensive Language Detection in comparison to the other models.

SaudiBERT [27] focuses on Saudi Arabian dialects, trained on a large corpus of Saudi Arabic text from Twitter and online sources. SaudiBERT has the same architecture as BERT but differs in the tokenizer, where the SentencePiece [43] tokenizer was used. The author compared the performance of SaudiBERT against AraBERTv02-Twitter, QARiB, CAMeLBERT-DA, MARBERTv1, MARBERTv2, and AraRoBERTa-SA on SA, Dialect Identification, Gender Identification, Event Detection, and Sarcasm Detection. The results showed that SaudiBERT outperformed the other models in all tasks except for one of the dialect identification tasks.

EgyBERT [29] specializes in Egyptian Arabic, addressing tasks such as gender identification and sarcasm detection. The model has the same architecture as BERT-Base, but similar to SaudiBERT, the SentencePiece tokenizer was used. The model was compared against AraBERTv02-Twitter, QARiB, CAMeLBERT-DA, MARBERTv1, and MARBERTv2 on Dialect, Abusive Language, Gender, and Sarcasm Detection, Sentiment and Emotion Analysis, and Topic Characteristic. The results show that EgyBERT performed best in most tasks, except for Gender Identification, Topic Characteristic, and one of the SA tasks where MARBERT performed better.

As we can see, the previously discussed models follow the BERT-based architecture with modifications in parameter size and maximum sequence length to accommodate dialect-specific features. They have shown superior performance in their respective dialects compared with general Arabic or multilingual models, particularly in tasks such as SA and Dialect Identification. However, it is important to note that the development of dialect-specific models is not uniform across the Arab world. While countries such as Sudan, Algeria, Tunisia, Morocco, Saudi Arabia, and Egypt have published works on their dialects, many other Arabic-speaking countries lack similar resources. This limitation points to a significant gap in the field where certain dialects and regional variations remain underrepresented in the current landscape of Arabic NLP models.

#### *5.1.3GPT-Based Models*

Following the emergence of encoder-only BERT-based models, decoder-only models, namely, GPT, have also been adapted for Arabic, offering new capabilities in text generation and completion tasks. These models have shown promise in various applications ranging from question-answering to language understanding.

The first published work that leveraged the GPT-2 architecture for Arabic is AraGPT2 [34], with four size variants: base (135 million parameters), medium (370 million parameters), large (792 million parameters), and mega (1.46 billion parameters), noting that the large and mega variants utilized the GROVER model [62]. All variants were trained on a substantial dataset comprising of 8.8 billion words. The perplexity score, which measures the degree of uncertainty of the model when assigning probabilities to the text, was used to evaluate the model. In zero-shot QA and translation downstream tasks, AraGPT2 performed well in zero-shot QA related to countries, birth and death years, and geography but struggled with questions involving quantities. The model's performance in English-to-Arabic translation was poor. Notably, a human evaluation revealed that AraGPT-2-mega could fool 60% of the subjects, with longer passages being more convincing than shorter ones.

Building upon AraGPT2, AraQA [48] was developed as an Arabic generative question-answering model for religious Arabic text, which was fine-tuned using the AraGPT2 architecture on 88.6 thousand question-answer pairs. Although the authors evaluated the model using perplexity and cross-entropy loss, they did not compare it to existing Arabic models or benchmarks, limiting the context for its performance.

JASMINE [35], based on the GPT-3 architecture, represents a significant scaling up of Arabic language models, with four size variants and a number of parameters varying between 350 million and 6.7 billion parameters. The dataset that the model was trained on comprises of 71.5 billion tokens of classical, modern, and dialectical Arabic. Two evaluation strategies were followed: Intrinsic Evaluation where the Perplexity metric was used, and Extrinsic Evaluation, where zero-shot, one-shot, and few-shots were used to evaluate the model on five tasks: language modeling, autocompletion, common sense inference, word manipulation, and natural language understanding. The authors compared JASMINE with AraGPT2 and mGPT for the perplexity score, and the results showed that JASMINE-6.7B obtained the best score. As for the downstream task results with different shot settings, it was concluded that JASMINE-6.7B performed best in autocompletion, word manipulation, and natural language understanding, whereas JASMINE-2.7B performed best in common sense inference.

ArabianGPT [49] has two variants: ArabianGPT-0.1B (small-scale) and ArabianGPT-0.3B (medium-scale), following the GPT-1 and GPT-2 medium architectures, respectively. The model was evaluated using a few-shot framework testing scientific reasoning, common sense understanding, cross-domain knowledge, and truthfulness, and was compared to the AraGPT-Base, AraGPT-Medium, Bloom-7b1, and Llama-7B models. The results show that ArabianGPT, in both size variants, excels in truthfulness, whereas the multilingual models (Bloom-7b1 and Llama-7B), performed better than ArabianGPT and AraGPT in the other tasks.

GPT-based models, built on the foundation of BERT-based models, have introduced significant advancements in Arabic NLP. One of the key differences is the dramatic increase in the parameter count. While BERT-based models typically range from tens to hundreds of millions of parameters, GPT-based models such as JASMINE scale up to 6.7 billion parameters, allowing for greater language understanding and generation capabilities. In addition, GPT models have shifted the focus of evaluation metrics. Instead of relying solely on task-specific metrics, such as accuracy and F1 score, GPT models use perplexity as a key intrinsic metric, offering a more comprehensive evaluation of language modeling abilities across various tasks. Moreover, GPT-based models excel in generative tasks such as zero-shot learning, text generation, and question answering. Notably, Arabic-specific models such as Arabian GPT often outperform multilingual models in tasks requiring deep cultural and linguistic understanding.

#### *5.1.4Other Models*

While BERT-based models dominate the landscape of Arabic language models, several other architectures have been explored to push the boundaries of Arabic language models. These include models based on ELECTRA, T5, BART, and RoBERTa, each of which has unique strengths in the field.

AraELECTRA [15] was the first Arabic language model based on the ELECTRA architecture, which is larger than BERT, consisting of 136 million parameters. The model was trained on a dataset of 8.8 billion words and evaluated on three downstream tasks: QA, SA, and NER. The authors compared their model against AraBERT, Arabic-BERT, and ARBERT. The evaluation results showed that AraELECTRA achieved the highest performance in all tasks, except for one QA task, where the much larger AraBERTv0.2-large maintained an edge.

AraBART [24], which utilizes the BART-Base architecture, with 139M parameters, was trained using the same corpus as AraBERT. The model focused on abstractive summarization and was compared to the monolingual Bert2Bert model, called C2C [64], multilingual mBART25 [65], and mT5 [66] models. The results showed that AraBART surpassed C2C across all evaluated summarization tasks, whereas mBART25 performed better in some instances.

AraMUS [46] is a larger encoder-decoder model based on the T5-xxl architecture with 11 billion parameters trained on 529 GB of Arabic text. AraMUS was fine-tuned on the ALUE benchmark, which includes eight tasks: Sentiment Intensity Regression, Emotion Classification, Sentence Classification, Sematic Question Similarity, Question Answering, Question Generation, and Text Summarization. Compared to ARBERT, MARBERT, JABER, SABER, ALM-1.0, AT5B and AraT5-base, AraMUS showed significant performance in all tasks.

AraStories [50] is a story generation model fine-tuned using LLaMA2, exploring two fine-tuning strategies: fine-tuning on data translated from English to Arabic, and fine-tuning data produced by GPT-4 in three Arabic dialects: MSA, Moroccan, and Egyptian. Based on the fine-tuning strategies, two models were generated: Model A, which followed the supervised fine-tuning approach on data generated from GPT-4, and Model B, which followed a two-step fine-tuning approach, where the model was first trained on translated data from English and then further instruct fine-tuning the model using data generated from GPT-4. The authors used GPT-4 as an evaluator for model performance as well as for human evaluation. Both models were compared to GPT-3.5, Command-R, and AceGPT-Chat for all Arabic varieties. The GPT-4 evaluation results showed that in the MSA form, Command-R had the best fluency, coherence, and instruction-following scores, whereas Model B had the best consistency score. Model B had the best fluency score in the Egyptian dialect and best consistency in the Moroccan dialect. In conclusion, Model B, which was exposed to more training data, performed better than Model A, and Command-R, which was much larger and trained on more data, showed strong performance across most metrics, outperforming GPT-3.5. As for the human evaluation, Model A tended to generate longer stories than Model B, and both models performed poorly on the Moroccan dialect compared to the Egyptian dialect, and both models outperform Command-R in the dialects. It is worth noting that GPT-3.5 and AceGPT-Chat failed to generate dialectal content.

Beyond BERT-based models, several other architectures have pushed the boundaries of Arabic language models, including ELECTRA, T5, BART, and RoBERTa. One particular model is AraMUS, which was the largest model among them at 11 billion parameters and showed significant gains across all evaluated tasks. Other architectures make different trade-offs, such as AraBART's focus on abstractive summarization and AraStories for story generation. Task-specific tuning strategies like these can yield customized models without the need for massive scale. This range of approaches beyond BERT-based pre-training highlights the nascent but rapid progress in Arabic language modeling.

### 5.2 Bilingual Models

Along with monolingual Arabic models, the field has witnessed the emergence of bilingual language models specifically trained on parallel corpora comprising both Arabic and English data. These bilingual Arabic-English models leverage the strengths and linguistic characteristics of both languages, enabling enhanced cross-lingual capabilities. This section delves into the architectural approaches, training methodologies, and evaluation benchmarks employed in developing bilingual Arabic-English language models, shedding light on their ability to bridge the linguistic and cultural gaps between these two widely spoken languages.

GigaBERT [17] is a BERT-based bilingual model designed for Information Extraction tasks in English and Arabic that focuses on NER, POS Tagging, Argument Labeling, and Relation Extraction. It comes in five variants, each trained on progressively larger datasets of English and Arabic data. Compared to baseline Arabic and multilingual models such as AraBERT, mBERT, XLM-R, and GigaXLM-R, GigaBERT variants demonstrated superior performance in most tasks across English, Arabic, and zero-shot transfer scenarios. However, XLM-R outperformed GigaBERT in POS tagging and Argument Labeling for English and zero-shot transfer tasks.

Jais [36] is a 13B parameter Arabic-English decoder-only language model based on the GPT-3 architecture, trained on 348B tokens (116B Arabic, 232B English), with the Arabic dataset augmented from an initial 55B tokens. Its instruction-tuned variant, Jais-chat, was fine-tuned on 9.6M instruction-response pairs. Both models were evaluated against baselines like AraT5, AraBART, mT0, BLOOM, BLOOMz, LLaMA, LLaMA2, and LLaMA2-chat on tasks including World Knowledge, Commonsense Reasoning, and Misinformation and Bias in Arabic and English. Jais and Jais-chat outperformed these baselines, establishing themselves as state-of-the-art for Arabic LLMs, particularly in terms of knowledge acquisition and commonsense reasoning. BLOOMz emerged as the best baseline for Arabic, while instruction-tuning further improved performance, with Jais-chat achieving the highest scores. Remarkably, Jais-chat excelled even in English tasks, surpassing dedicated English models despite having less training data and a smaller parameter size.

AceGPT [39] is an LLaMA2-based model, with AceGPT-chat being a version that has undergone supervised fine-tuning and reinforcement learning from AI feedback. The model was trained using two versions of LLaMA2: LLaMA2-7B and LLaMA2-13B. The 7B version was trained with 30B tokens (19.2B in Arabic and 10.8B in English), while the 13B version was trained with 10B tokens (6B in Arabic and 4B in English). The baseline models used for comparison included LLaMA2, Bloomz, GPT-3.5 Turbo, and Jais. The results showed that the AceGPT models excel in instruction-following tasks, human evaluation, and knowledge-based tasks. While GPT-3.5 Turbo achieved strong results in various tasks, AceGPT-13B-chat performed competitively, particularly in areas related to Arabic language and culture.

ALLaM [51] is a series of large language models developed by the National Center for AI in Saudi Arabia to advance Arabic language technologies. The models include a 7B parameter version trained from scratch, as well as 13B and 70B parameter versions initialized from LLaMA-2 weights. The pretraining dataset comprised 4T English tokens and 540B Arabic tokens, including both the original Arabic text and machine-translated English data. The training methodology utilized tokenizer augmentation to handle Arabic, initialized embeddings for new Arabic vocabulary, and a two-step pre-training recipe. First, the 7B model was trained on four trillion English tokens, then both the 7B and larger models were trained on a mixture of 1.2T English and Arabic tokens. The ALLaM models were evaluated on benchmarks covering capabilities such as reasoning, world knowledge, language understanding, safety, conversation, math, and coding. For Arabic, ALLaM-70B achieved state-of-the-art results in most tasks. In English evaluations, ALLaM-70B performed the second-best overall behind LLaMA-3.

Following the release of ALLaM, Stability AI released Arabic Stable LM [45], a 1.6 billion parameter model derived from Stable LM 2 and fine-tuned on over 100 billion Arabic tokens. The authors intentionally maintained a smaller parameter count to optimize for efficiency while maximizing Arabic language capabilities. The model was released in two variants: a base model and a chat model. The evaluation process incorporated multiple Arabic benchmarks to assess both general language understanding and cultural alignment. Notably, the Arabic Stable LM chat variant outperformed larger Arabic language models, including Jais and AceGPT, across these benchmarks despite its comparatively smaller parameter count.

One challenge in Arabic LLMs is the lack of data which led to slow computation speed in encoding and decoding the Arabic language in large-scale multilingual LLMs, such as LLaMA and GPT. This challenge led to the development of AraLLaMA [52], a fine-tuned Arabic LLM based on the LLaMA2 architecture with 13B parameter size, that employs progressive vocabulary expansion, implemented by a modified Byte Pair Encoding (BPE) algorithm that progressively extends the Arabic subwords in its dynamic vocabulary during training. The model was evaluated on world knowledge, reading comprehension, question answering, and cultural alignment and achieved impressive results compared to larger Arabic models, such as AceGPT and Jais.

The latest addition to Arabic LLMs, as of February 2025, is Fanar [41], an Arabic multimodal LLM developed at Hamad Bin Khalifa University's Qatar Computing Research Institute (QCRI) with two variants: Fanar Star, a 7B parameter model trained from scratch using the OLMo and Llama architectures, and Fanar Prime, a 9B parameter model trained on the Gemma-2 base model. The pretraining dataset comprised of 513B English tokens and 410B Arabic tokens. Fanar models were benchmarked on open-ended and conversational assessments, human evaluation, and automatic evaluation on downstream tasks, such as cultural alignment, dialects identification, and machine translation tasks. The results show that Fanar Prime achieves the best accuracy in most of the benchmarks, particularly cultural assessments and dialectal tasks, and offers competitive performance in generative open-ended tasks judged by LLMs and humans. In addition to the Fanar's capabilities in language processing tasks, it also provides multimodal support for speech and image generation tasks. Also, it provides capability in RAG systems for Islamic content, recent information, content attribution, and select biographies.

### 5.3 Multilingual Models

In recent years, there has been a surge in the development of LLMs; however, most of them are multilingual, supporting multiple languages, including Arabic. ArabicBERT [18] is an Arabic model that uses a combination of BERT and Convolution Neural Network (CNN) layers. The model was trained on three languages: Arabic, Greek, and English. ArabicBERT has four variants of different sizes trained on the same data. The ArabicBERT model was compared to SVM with TF-IDF, multilingual BERT, Bi-LSTM, CNN-Text, and BERT, for Offensive Language Detection. The results show that ArabicBERT outperformed other baseline models in all languages except Turkish, where BERT performed the best.

AraT5 [47] was the first sequence-to-sequence Arabic language model using a T5-Base encoder-decoder architecture. AraT5 has three models, each of which is trained on a different type of data. The first model is AraT5-MSA, trained on MSA data; the second model is AraT5-TW, trained on Twitter data; and lastly, AraT5, which was trained on both MSA and Twitter data. The authors compared their models to a vanilla sequence-to-sequence transformer (S2S) [67] and mT5 on the seven downstream tasks, namely: MT, code-switched translation (CST), text summarization (TS), news title generation (NGT), question generation (QG), transliteration (TR), and paraphrasing (PPH). The results show that AraT5-MSA performed best in all tasks, outperforming multilingual models with five times more data size. Although AraT5 trained specifically to perform tasks in the Arabic language, the training data contained vocabulary from 11 languages other than Arabic, and hence can be further trained/fine-tuned in other languages.

Other large-scale multilingual language models trained on Arabic data include Aya [68], Command-R [69], BLOOM [70], and Qwen [71]. Aya and Command-R were developed by Cohere [72], where Aya is an instruction-finetuned language model based on the mT5 baseline model in 101 languages and Command-R is a generative model for retrieval-augmented generation (RAG) tasks comprising of 35B parameters and trained on 24 languages. BLOOM comprises of 176B parameters and is trained on 46 languages. Qwen is an instruction-tuned LLM built by Alibaba Cloud [73] with different size variants from 0.5B until 72B parameters. Additionally, there have been efforts by companies such as Google and Meta to develop LLMs that support a wide variety of languages, namely Gemini [74] and Llama. These models, while not exclusively focused on Arabic, are trained on substantial Arabic data and can be adapted for Arabic NLP tasks through fine-tuning or prompting. It is worth noting that the development of Arabic language models is still in its early stages compared to English and other well-resourced languages. However, the increasing availability of Arabic data and the advancements in language modeling techniques have led to significant progress in this area, with more sophisticated and capable Arabic LLMs expected to emerge in the coming years.

### 5.4 Discussion

Figure 5 illustrates the geographic distribution of Arabic LLMs, showing how different models are trained and developed across regions, offering insight into regional disparities and concentrations in Arabic language model development.

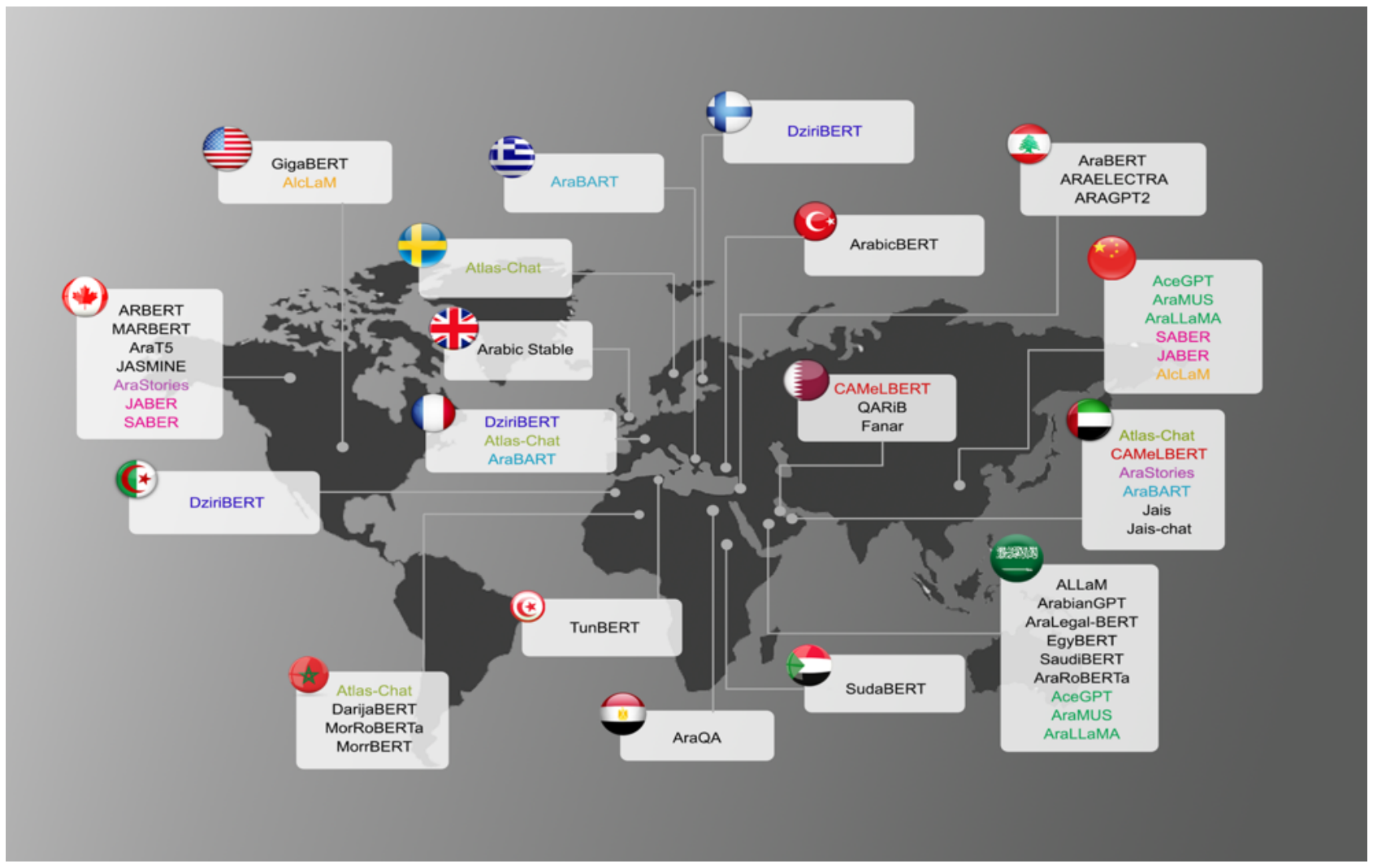

Figure 5: Geographic Distribution and Development of Arabic LLMs. Model names with the same color indicate collaborative development efforts between different countries.

Based on our previous overview of Arabic LLMs, the landscape of Arabic language models encompasses a diverse range of architectures tailored to different use cases and capabilities. As summarized in Table 7, monolingual models focus exclusively on the Arabic language, allowing a nuanced understanding of dialects, syntax, and other linguistic features.

Architecturally, BERT-based encoders continue to dominate, but incremental innovations with models such as AraGPT, AraELECTRA, and AraMUS have brought new capabilities, establishing architectures such as GPT, ELECTRA, and T5. Various capabilities and downstream tasks have been tackled by Arabic LLMs, as summarized in Figure 6, where Arabic LLMs' performance has surged on tasks such as SA, NER, QA, and dialect identification, often approaching or exceeding multilingual baselines. Evaluation metrics vary widely across models, with accuracy and F1 score emerging as the most used metrics. Addressing these gaps presents an opportunity to widen accessibility and enhance the understanding of Arabic language modeling struggles. With computing resources expansion and accelerating model scaling, the future is bright for monolingual Arabic models that address a wide range of applications.

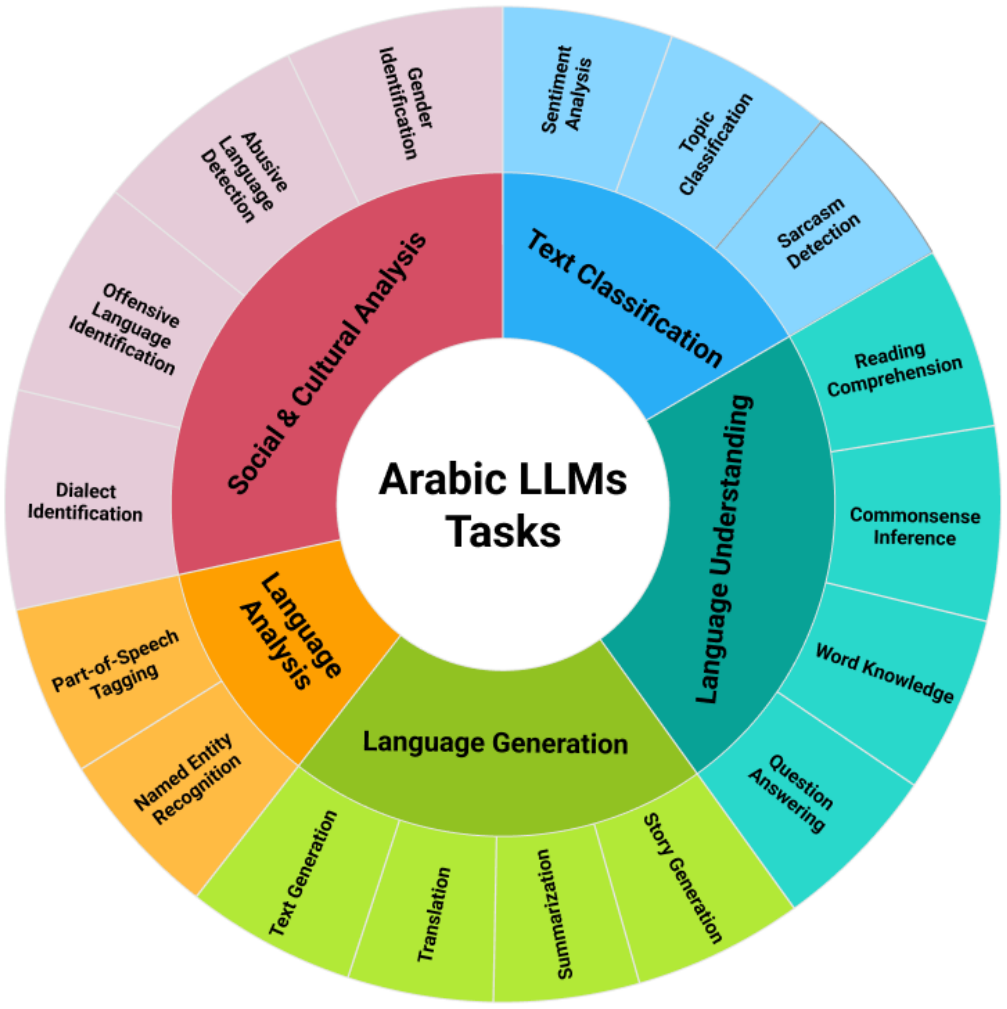

Figure 6: Overview of Downstream Tasks tackled by Arabic LLMs

In contrast, bilingual Arabic-English models do not rely primarily on BERT-based architectures. Instead, they leverage other established architectures such as GPT and LLaMA as their foundations. For example, as summarized in Table 8, Jais utilizes the GPT-3 architecture, and its instruction-tuned variant Jais-chat builds upon this decoder-only foundation. Similarly, AceGPT employs the LLaMA2 architecture. This highlights the architectural diversification in bilingual modeling beyond BERT encoders. Bilingual models also operate at substantially larger scales than their monolingual counterparts do. For instance, Jais comprises 13B parameters trained on 348B tokens, whereas ALLaM scales up to 70B parameters trained on 4T English and 540B Arabic tokens. The data scale surpassed most monolingual models by an order of magnitude. AceGPT likewise trains its 7B parameter version on 30B tokens. This demonstrates the need for massive amounts of data to train capable bilingual models. As evidenced in downstream evaluations, they achieved state-of-the-art capabilities in knowledge-intensive domains, such as reasoning, linguistics, and common sense tasks across English and Arabic. However, their extensive data dependence also poses challenges regarding equitable Arabic representations and cultural grounding.

Finally, multilingual models (summarized in Table 9 provide general linguistic capabilities across dozens of languages, including Arabic. While building on established LLMs, such as T5 and BERT, these models show strong cross-lingual transfer but lag behind specialized models in Arabic-specific understanding.

Table 7: Summary of monolingual language models for Arabic and its dialects, including variants, number of parameters, vocabulary size, context/sequence size, downstream tasks, hardware, training time, and evaluation metrics. Data as of February 2025

| Model | Variants | # Parameters | Vocabulary Size | Context/Sequence Size | Downstream Task | Hardware | Training Time | Evaluation Metrics |
|---|---|---|---|---|---|---|---|---|
| AraBERT [2] | AraBERT-v0.1 AraBERT-v1 | 110M | 64K | 128 for the first 900K steps, 512 for the remaining steps | SA, NER, QA | TPUv2-8 | 4 days | Accuracy and macro F1 Score |
| MARBERT [4] | - | 163M | 100K | 128 | SA, Social Meaning, Topic Classification, Dialect Identification, NER | 8 Google Cloud TPUs | 40 days | Accuracy and F1 Score |
| ARBERT [4] | - | 163M | 100K | 128 | | 1 Google Cloud TPU with 8 cores (v2.8) | 16 days | |
| QARiB [3] | - | 110M | N/A | 128 | Text Categorization | Google TPU | - | F1 Score |
| SudaBERT [20] | - | 110M | N/A | 128 | SA, NER | Google TPU (v3-8) | 14 hours | Accuracy and F1 Score |
| AraELECTRA [19] | - | 136M | N/A | 384 (for QA), 256 (for SA and NER) | Reading comprehension, SA, NER | Google TPU (v3-8) | 24 days | Exact Match and F1 Score |
| AraGPT2 [34] | Base, Medium, Large, Mega | 135M, 370M, 792M, 1.46B | 64K | 1024 | Zero-shot QA, English-to-Arabic Translation | TPUv3-128 Slice | 1.5 days (Base), 23 days (Medium), 3 days (Large), 9 days (Mega) | Perplexity and Human Evaluation |
| CAMeLBERT [19] | CAMeLBERT-MSA, CAMeLBERT-DA, CAMeLBERT-CA, CAMeLBERT-Mix | N/A | 30K | 128 for the first 900K steps, 512 for the remaining 100K steps | NER, POS tagging, SA, Dialect Identification, Poetry Classification | Google Cloud TPU (v3-8) | 4.5 days | F1 Score and Accuracy for POS tagging only |
| JABER [22] | - | 135M | 64K | 128 | 4 single-sentence, 2 sentence-pair, one multi-label classification, single regression | 8 NVIDIA Tesla V100 GPUs | 10 days | Standard Deviation |
| SABER [22] | - | 369M | | | | | 6.7 days | |
| AraBART [24] | - | 139M | 50K | N/A | Abstractive Summarization | 128 Nvidia V100 GPUs | 60 hours | F1 Score |
| AraLegal-BERT [21] | AraLegal-BERT-Base, AraLegal-BERT-Large | 110M | 64K | 512 | Legal text classification, Keyword Extraction, NER | 8 NVIDIA DGX1 GPUs | N/A | Precision, Recall, and F1 Score |
| AraRoBERTa [20] | AraRoBERTa-SA, AraRoBERTa-EG, AraRoBERTa-KU, AraRoBERTa-OM, AraRoBERTa-LB, AraRoBERTa-JO, AraRoBERTa-DZ | 126M | 52K | 512 | Text Classification | 1 16GB NVIDIA Tesla P100 GPU | N/A | F1 Score |
| DziriBERT [23] | - | 124M | 50K | 64 | SA, Emotion Classification, Topic Classification | 1 NVIDIA T4 GPU | 10 days | Accuracy, F1 Score, Precision, and Recall |
| TunBERT [25] | TunBERT-T, TunBERT-P | 110M | 48.2K | 128 | SA, Dialect Identification, Reading Comprehension, QA | 4 NVIDIA Tesla V100 GPUs | 5 days | Accuracy and F1 Score |
| DarijaBERT [31] | DarijaBERT, DarijaBERT-arabizi, DarijaBERT-mix | 147M, 170M, 209M | 80K, 110K, 160K | 128 | Dialect identification, SA, Sarcasm Detection, Topic Classification | 1 Google TPU (v3.8) | 49 hours, 60 hours, 96 hours | F1 Score and Accuracy |
| AraMUS [46] | - | 11B | 64K | N/A | Sentiment Intensity Regression, Emotion Classification, Sentence Classification, Sematic Question Similarity, QA, Question Generation, Text Summarization | 128 NVIDIA A100 GPUs | 2 months | Exact Match and F1 Score |
| MorRoBERTa [26] | - | 83M | 52K | 512 | SA, Dialect Identification, Language, Classification | HPC-MARWAN GPU | 92 hours | Accuracy and F1 Score |
| MorrBERT [26] | - | 125M | | | | | 120 hours | |
| JASMINE [35] | JASMINE350M, JASMINE1.3B, JASMINE2.7B, JASMINE6.7B. | 350M, 1.3B, 2.7B, 6.7B | 64K | 128 | Language modeling, Autocompletion, Commonsense inference, Word manipulation, Natural language understanding. | Google TPUs | N/A | Perplexity and F1 score |
| AraQA [48] | - | 135M | 64K | 1024 | QA | 1 NVIDIA GeForce RTX | 7 hours and 4 mins | Perplexity and cross-entropy loss function |
| | | | | | | 3070 8GB GDDR6 | | |
| ArabianGPT [49] | ArabianGPT-0.1B, ArabianGPT-0.3B | 0.1B (134M), 0.3B (345M) | 64K | 768, 1024 | SA, Summarization | 2/4 NVIDIA A100 GPUs | 3 days, 4.1 days | Accuracy and F1 Score |
| AraPoemBERT [30] | - | 110M | 50K | N/A | SA, Poetry meters and sub-meters classification, Poet's Gender classification, Verses' Rhymes Detection. | 1 GeForce RTX 4090 GPU | 142 hours | Accuracy, F1 Score, Precision, and Recall |
| SaudiBERT [27] | - | 143M | 75K | 128 | SA, Dialect Identification, Gender Identification, Event Detection, Sarcasm Detection | 2 GeForce RTX 4090 GPUs | 342 hours | F1 Score and Accuracy |
| AlcLaM [28] | - | 125M | 52K | 128 | SA, Offensive Language Detection, Dialect Identification | N/A | N/A | F1 Score and Accuracy |
| AraStories [50] | Model A, Model B | 7B | N/A | N/A | Story Generation | 1 NVIDIA A100 GPU | 5.5 hours, 17.5 hours | GPT-4 and Human Evaluation |
| EgyBERT [29] | - | 144M | 75K | 128 | Dialect Detection, Abusive Language Detection, Sarcasm Detection, Gender Identification, SA, Emotion Analysis | 2 GeForce RTX 4090 GPUs | 774 hours | F1 Score and Accuracy |
| Atlas-Chat [43] | Atlas-Chat-2B, Atlas-Chat-9B | 2B, 9B | N/A | 2048 | SA, Translation, Summarization | 8 Nvidia A100 80 GB GPUs | N/A | Accuracy (MCQs) and BLEU Score |

Table 8: Summary of Bilingual LLMs for the Arabic language and its dialects, including variants, number of parameters, vocabulary size, context/sequence size, downstream tasks, hardware, training time, and evaluation metrics. Data as of February 2025

| Model | Variants | # Parameters | Vocabulary Size | Context/Sequence Size | Downstream Tasks | Hardware | Training Time | Evaluation Metrics |
|---|---|---|---|---|---|---|---|---|
| GigaBERT [17] | GigaBERT-v0, GigaBERT-v1, GigaBERT-v2, GigaBERT-v3, GigaBERT-v4 | 110M | 50K | 128 (GigaBERT-v0/1/2), 512 (GigaBERT-v3/4) | NER, POS Argument role labeling, Relation Extraction | Google Cloud TPUs | N/A | Accuracy and F1 Score |
| Jais [36] | Jais, Jais-chat | 13B | 84K | 2048 | Word Knowledge, Commonsense Reasoning, Misinformation and bias | Condor Galaxy 1 (CG-1) AI supercomputer | N/A | Accuracy |
| AceGPT [39] | AceGPT-7B, AceGPT-13B | 7B, 13B | 32K | 2048 | Instruction Following, Natural Language Understanding Knowledge Localization | 24 Nvidia A100 80G GPUs | N/A | Accuracy and F1 Score |
| ALLaM [51] | ALLaM-7B, ALLaM-13B | 7B, 13B, 70B | N/A | 1024 | Arabic and English language Understanding | 128-1024 A100 GPUs. | 5M GPU hours | Accuracy, GPT-4, and Human Evaluation |
| Arabic Stable LM [45] | Arabic Stable LM-base, Arabic Stable LM-chat | 1.6B | 100K | N/A | QA, SA, Reading Comprehension, Rating Classification | 8 H100 GPUs | N/A | Accuracy |
| AraLLaMa [52] | AraLLaMA-base, AraLLaMA-chat | 7B, 13B | 12.8K | 4096 | QA, World Knowledge, Reading Comprehension, Cultural Alignment | 2,368 GPUs | N/A | Accuracy |
| Fanar [41] | Fanar Star, Fanar Prime | 7B, 9B | 76K, 128K | 4096 | QA, Reasoning, Reading Comprehension, Conversation, Instruction-Following | 168 NVIDIA H100 80GB SXM5 GPUs | N/A | Accuracy |

Table 9: Summary of Multilingual LLMs for the Arabic language and its dialects, including variants, number of parameters, vocabulary size, context/sequence size, downstream tasks, hardware, training time, and evaluation metrics. Data as of February 2025

| Model | Variants | # Parameters | Vocabulary Size | Context/Sequence Size | Downstream Task | Hardware | Training Time | Evaluation Metrics |
|---|---|---|---|---|---|---|---|---|
| ArabicBERT [18] | Mini, Base, Medium, Large | 110M | 32K | 64 | Offensive Language Identification | 1 Google TPU (v3.8) | N/A | F1 Score |
| AraT5 [47] | AraT5-MSA, AraT5-TW, AraT5 | 220M | N/A | 512 (except AraT5-TW which is 128) | Machine translation, code-switched translation, text summarization, news title generation, question generation, transliteration, paraphrasing. | 8 Google TPUs (v3.8) | 80 days | BLEU Score |

## 6 ARABIC LLMS OPENNESS

In recent years, there has been an increase in language models claiming to be open, but how open are they really? This question raises the demand for creating an openness assessment to explore different dimensions of the openness of language models. Based on the literature [75] and [76], free availability is not equal to openness and transparency. In addition, making the weights of the models available in the name of openness while maintaining the architecture and how the system is built under wraps aligns with the practice of openwashing [77]. This section discusses 12 elements of openness, including the availability of the code and weights, clear and scientific documentation, and access methods, based on the framework of Liesenfeld and Dingemanse [78], which analyzes European LLMs and their openness alignment with the EU AI Act [79]. The EU AI Act was chosen to validate the openness of Arabic LLMs because of the unavailability of rules and regulations for the openness of AI in Arabic-speaking countries.

Table 10 summarizes the level of openness for the assessed models, categorized as fully open (✓), partially open (~), or closed (✗). The table evaluates each model based on key openness elements, including the availability of code, weights, data, and other relevant factors. A detailed discussion of these openness elements follows, providing a comprehensive assessment of how well these models meet the transparency standards.

### 6.1 Key Elements of Openness

The framework follows the composite and graded approach, which entails that the framework comprises multiple elements for assessments, where each element is graded as open, partial, or closed, to provide a well-informed and systematic judgment of the openness of the models. This section discusses three dimensions–availability, documentation, and access–and each element under the dimension that is used for grading.

#### *6.1.1 Availability*

The availability dimension contains six elements: open code, data, weights, instruction-tuning (RL) data, RL weights, and license. The open code indicates the availability of source code for training, fine-tuning, and running the model [80]. We can see that 16 of the 36 models fully satisfy the open code element, whereas six models partially make their codes available. Although ARBERT and MARBERT come from the same paper, the authors only provided code for fine-tuning MARBERT for sentiment analysis. The same applies to JABER and SABER, where the authors provided a code for running and fine-tuning JABER only, owing to the large size of the SABER. Some authors have not provided the code but are willing to provide it per request, such as the JASMINE model.

The availability of the data element refers to the public release of datasets, which enables access and reuse [81]. We can see that 17 models have their training data fully available and 17 models have their data partially available, particularly their own collected data. The RL data element refers to the data used for instruction-tuning of the model. While they are not utilized for BERT-based models, GPT- and Llama-based models have utilized instruction-tuning, particularity AraStories, and Atlas-Chat, where they make the RL data fully available. However, AceGPT, JAIS, and ALLaM make their RL data partially available.

The availability of the model weights enables the reconstruction and production of the full model. The majority of published models on huggingface have model weights available for usage. Some models, namely QARiB and AraQA, have their source code fully available, but the weight of the model has not been published, indicating that users can retrain the model from scratch using their own datasets; however, they cannot replicate the exact performance and outputs of the original trained model. RL weights refer to the weights of the model after instruction-tuning. We can see that only JAIS and AceGPT made their instruction-tuned model weights available.

Despite the rapid growth of open language models, approximately 65% of models on the Hugging Face are without a license [82]. Most models do not have software licenses. However, we can see that 16 out of the 36 models have a license for their models to be used, modified, and shared under defined terms and conditions.

### *6.1.2 Documentation*

The documentation dimension contains four elements: code, architecture, paper, and model card. The documentation of the code aligns with the availability of the *open code* element, such that when the code is available, it is most likely well-documented with the addition of comments and explanations to the source code. However, in some models, such as AraQA, AraPoemBERT, and SaudiBERT, the code is fully available, but the code documentation lacks some details. The JAIS and ArabicBERT models have their source code fully available but with no code documentation, making code understanding challenging for users.

The documentation of the models' *architecture* is usually fully described in the paper. Most of the models provide a full description of the model's architecture, including the base model, number of parameters, and layers. However, some models such as ARBERT, QARiB, and SudaBERT lack architectural details such as vocabulary and context size.

The documentation of models through research papers and model cards achieves full transparency and extensibility [80]. Most models have full details documented in research papers; many are missing the model card document. Only 12 out of 36 models had their model cards fully documented, reflecting a gap in the standardization of model documentation. The absence of model cards for many models suggests a need for improved documentation practices to enhance user understanding and transparency.

### *6.1.3 Access*

The access dimension includes two elements: access via an API and downloadable software package. Most models are available in HuggingFace and can be accessed through a pipeline API call for model usage or an installed library package via a public code repository from GitHub. The only closed-source models with no available access are SudaBERT, SABER, AraLegal-BERT, AraQA, JASMINE, AraMUS, and AraQA-BERT, which are proprietary and restricted due to their specific use cases or the preferences of their developers.

Table 10: Openness levels of monolingual and bilingual Arabic LLMs, including availability of code, data, weights, RL elements, and documentation

| Type | Model | Availability | | | | | Documentation | | | | | Access | |
|---|---|---|---|---|---|---|---|---|---|---|---|---|---|
| | | Open Code | Data | Weights | RL Data | RL Weights | License | Code | Architecture | Paper | Model Card | API | Package |
| Monolingual | AraBERT | ✓ | ~ | ✓ | - | - | ✓ | ✓ | ✓ | ✓ | ✓ | ✓ | ✓ |
| | MARBERT | ~ | ~ | ✓ | - | - | ✗ | ~ | ✓ | ✓ | ✗ | ✓ | ✓ |
| | ARBERT | ✗ | ~ | ✓ | - | - | ✗ | ✗ | ~ | ✓ | ✗ | ✓ | ✓ |
| | QARiB | ✓ | ✓ | ✗ | - | - | ✓ | ✓ | ~ | ✓ | ✓ | ✗ | ✓ |
| | SudaBERT | ✗ | ~ | ✗ | - | - | ✗ | ✗ | ~ | ✓ | ✗ | ✗ | ✗ |
| | AraELECTRA | ✓ | ~ | ✓ | - | - | ✓ | ✓ | ✓ | ✓ | ✓ | ✗ | ✓ |
| | AraGPT2 | ✓ | ~ | ✓ | - | - | ✓ | ✓ | ✓ | ✓ | ✓ | ✗ | ✓ |
| | CAMeLBERT | ✓ | ✓ | ✓ | - | - | ✓ | ✓ | ✓ | ✓ | ✓ | ✗ | ✓ |
| | JABER | ~ | ✓ | ✓ | - | - | ✓ | ~ | ✓ | ✓ | ✗ | ✗ | ✓ |
| | SABER | ✗ | ✓ | ✗ | - | - | ✗ | ✗ | ✓ | ✓ | ✗ | ✗ | ✗ |
| | AraBART | ✗ | ✓ | ✗ | - | - | ✗ | ✗ | ✓ | ✓ | ✗ | | ✓ |
| | AraLegal-BERT | ✗ | ~ | ✗ | - | - | ✗ | ✗ | ✓ | ✓ | ✗ | ✗ | ✗ |
| | AraRoBERTa | ✗ | ~ | ✗ | - | - | ✗ | ✗ | ✓ | ✓ | ✗ | ✗ | ✓ |
| | DziriBERT | ✓ | ~ | ✓ | - | - | ✓ | ✓ | ✓ | ✓ | ✗ | ✗ | ✓ |
| | TunBERT | ✓ | ✓ | ✓ | - | - | ✓ | ✓ | ✓ | ✓ | ✗ | ✗ | ✓ |
| | AraQA | ✓ | ✓ | ✗ | - | - | ✗ | ~ | ✓ | ✓ | ✗ | ✗ | ✗ |
| | JASMINE | ✗ | ~ | ✗ | - | - | ✗ | ✗ | ✓ | ✓ | ✗ | ✗ | ✗ |
| | AraMUS | ✗ | ~ | ✗ | - | - | ✗ | ✗ | ~ | ✓ | ✗ | ✗ | ✗ |
| | DarijaBERT | ~ | ✓ | ✓ | - | - | ✗ | ~ | ✓ | ✓ | ~ | ✗ | ✓ |
| | MorrBERT | ✗ | ~ | ✓ | - | - | ✗ | ✗ | ✓ | ✓ | ✗ | ✓ | ✓ |
| | MorRoBERTa | ✗ | ~ | ✓ | - | - | ✗ | ✗ | ✓ | ✓ | ✗ | ✓ | ✓ |
| | ArabianGPT | ✗ | ~ | ✓ | - | - | ✗ | ✗ | ✓ | ✓ | ✓ | ✗ | ✓ |
| | AraPoemBERT | ✓ | ✓ | ✓ | - | - | ✗ | ~ | ✓ | ✓ | ✗ | ✗ | ✓ |
| | SaudiBERT | ✓ | ✓ | ✓ | - | - | ✗ | ~ | ✓ | ✓ | ✗ | ✗ | ✓ |
| | AlcLaM | ✓ | ~ | ✓ | - | - | ✓ | ~ | ✓ | ✓ | ✗ | ✗ | ✓ |
| | AraStories | ~ | ✓ | ✗ | ✓ | ✗ | ✓ | ~ | ~ | ✓ | ✗ | ✗ | ✓ |
| | EgyBERT | ✓ | ✓ | ✓ | - | - | ✓ | ~ | ✓ | ✓ | ✗ | ✗ | ✓ |
| | Atlas-Chat | ~ | ✓ | ✓ | ✓ | ✗ | ✗ | ✗ | ✓ | ✓ | ✓ | ✓ | ✓ |
| Bilingual | GigaBERT | ✓ | ✓ | ✓ | - | - | ✓ | ~ | ✓ | ✓ | ✓ | ✓ | ✓ |
| | Jais | ✓ | ~ | ✓ | ~ | ✓ | ✓ | ✗ | ✓ | ✓ | ✓ | ✓ | ✓ |
| | AceGPT | ~ | ✓ | ✓ | ~ | ✓ | ✓ | ~ | ~ | ✓ | ✓ | ✓ | ✓ |
| | ALLaM | ✗ | ~ | ✗ | ~ | ✓ | ✓ | ✗ | ~ | ✓ | ✓ | ✓ | ✓ |
| | Arabic Stable LM | ✗ | ✓ | ✓ | ✓ | ✓ | ✓ | ✗ | ✓ | ✓ | ✓ | ✓ | ✓ |
| | AraLLaMa | ✗ | ✓ | ✗ | ✓ | ✗ | ✗ | ✗ | ~ | ✓ | ✗ | ✗ | ✗ |
| | Fanar | ✗ | ~ | ✗ | ~ | ✗ | ✗ | ✗ | ✓ | ✓ | ✗ | ✓ | ✗ |
| Multilingual | ArabicBERT | ✓ | ✓ | ✓ | - | - | ✓ | ✗ | ✓ | ✓ | ✗ | ✓ | ✓ |
| | AraT5 | ✓ | ~ | ✓ | - | - | ✗ | ~ | ~ | ✓ | ~ | ✓ | ✓ |

## 7 CONCLUSION

This survey provides a comprehensive overview of the current state of LLMs in Arabic NLP. We explored various aspects of Arabic LLMs including their architecture, training datasets, evaluation metrics, and openness.

Our analysis of the datasets used for pre-training Arabic LLMs highlights the diversity of available resources, including CA, MSA, and DA. We found that, while MSA resources are relatively abundant, there is a need for more datasets representing CA and a wide range of Arabic dialects. Addressing this data imbalance is crucial for the development of more inclusive and representative Arabic LLMs.

We have also delved into the realm of monolingual, bilingual, and multilingual LLMs for the Arabic language and its dialects. Our review of these models shows the variety of architectures employed, including encoder-only models (e.g., BERT, ELECTRA, and RoBERTa), decoder-only (e.g., GPT), and encoder-decoder models (e.g., T5). These LLMs have been applied to a wide range of downstream tasks such as sentiment analysis, named entity recognition, question answering, and machine translation. Evaluations against baseline models demonstrated the significant progress made in Arabic NLP using LLMs.

Furthermore, we assessed the openness of Arabic LLMs using a comprehensive framework that considered factors such as the availability of source code, training data, model weights, and documentation. Our findings highlight the importance of openness for reproducibility and transparency and facilitate further research and development in the field.

Despite the advancements in Arabic LLMs, there are still several challenges and opportunities for future research. A key direction is the development of more diverse and representative datasets, particularly for CA and dialectal variations. This will enable the creation of LLMs that can better capture the richness and nuances of the Arabic language across its various forms.

Another important avenue for future research is the exploration of novel architectures and training techniques specifically tailored to the unique characteristics of Arabic such as its complex morphology and syntax. This could involve the development of Arabic-specific tokenization methods, attention mechanisms, or pre-training objectives that consider the linguistic properties of a language.

In addition, there is a need for more standardized evaluation benchmarks and metrics for Arabic LLMs. The establishment of common datasets and tasks will facilitate the comparison and assessment of different models, thereby driving further progress in the field.

In conclusion, this survey provides a valuable resource for researchers, practitioners, and enthusiasts interested in the current state and future direction of Arabic LLMs. By highlighting the achievements, challenges, and opportunities in this field, we hope to inspire further research and development efforts to advance Arabic NLP and unlock the full potential of language technologies for Arabic-speaking communities worldwide.

## REFERENCES

[1] O. ElJundi, W. Antoun, N. El Droubi, H. Hajj, W. El-Hajj, and K. Shaban, "hULMonA: The Universal Language Model in Arabic," in *Proceedings of the Fourth Arabic Natural Language Processing Workshop*, W. El-Hajj, L. H. Belguith, F. Bougares, W. Magdy, I. Zitouni, N. Tomeh, M. El-Haj, and W. Zaghouani, Eds., Florence, Italy: Association for Computational Linguistics, Aug. 2019, pp. 68–77. doi: 10.18653/v1/W19-4608.

[2] W. Antoun, F. Baly, and H. Hajj, "AraBERT: Transformer-based Model for Arabic Language Understanding," in *Proceedings of the 4th Workshop on Open-Source Arabic Corpora and Processing Tools, with a Shared Task on Offensive Language Detection*, H. Al-Khalifa, W. Magdy, K. Darwish, T. Elsayed, and H. Mubarak, Eds., Marseille, France: European Language Resource Association, May 2020, pp. 9–15. Accessed: May 09, 2024. [Online]. Available: https://aclanthology.org/2020.osact-1.2

[3] S. A. Chowdhury, A. Abdelali, K. Darwish, J. Soon-Gyo, J. Salminen, and B. J. Jansen, “Qarib,” in *Proceedings of the Fifth Arabic Natural Language Processing Workshop*, I. Zitouni, M. Abdul-Mageed, H. Bouamor, F. Bougares, M. El-Haj, N. Tomeh, and W. Zaghouani, Eds., Barcelona, Spain (Online): Association for Computational Linguistics, Dec. 2020, pp. 226–236. Accessed: Aug. 28, 2024. [Online]. Available: https://aclanthology.org/2020.wanlp-1.21

[4] M. Abdul-Mageed, A. Elmadany, and E. M. B. Nagoudi, “ARBERT & MARBERT: Deep Bidirectional Transformers for Arabic,” in *Proceedings of the 59th Annual Meeting of the Association for Computational Linguistics and the 11th International Joint Conference on Natural Language Processing (Volume 1: Long Papers)*, C. Zong, F. Xia, W. Li, and R. Navigli, Eds., Online: Association for Computational Linguistics, Aug. 2021, pp. 7088–7105. doi: 10.18653/v1/2021.acl-long.551.

[5] J. Devlin, M.-W. Chang, K. Lee, and K. Toutanova, “BERT: Pre-training of Deep Bidirectional Transformers for Language Understanding,” May 24, 2019, *arXiv*: arXiv:1810.04805. doi: 10.48550/arXiv.1810.04805.

[6] K. Clark, M.-T. Luong, Q. V. Le, and C. D. Manning, “ELECTRA: Pre-training Text Encoders as Discriminators Rather Than Generators,” Mar. 23, 2020, *arXiv*: arXiv:2003.10555. doi: 10.48550/arXiv.2003.10555.

[7] Y. Liu *et al.*, “RoBERTa: A Robustly Optimized BERT Pretraining Approach,” Jul. 26, 2019, *arXiv*: arXiv:1907.11692. Accessed: Sep. 16, 2024. [Online]. Available: http://arxiv.org/abs/1907.11692

[8] A. Radford, J. Wu, R. Child, D. Luan, D. Amodei, and I. Sutskever, “Language Models are Unsupervised Multitask Learners”.

[9] C. Raffel *et al.*, “Exploring the Limits of Transfer Learning with a Unified Text-to-Text Transformer,” Sep. 19, 2023, *arXiv*: arXiv:1910.10683. doi: 10.48550/arXiv.1910.10683.

[10] “NOOR, the new largest NLP model for the Arabic language | MultiLingual.” Accessed: Oct. 07, 2024. [Online]. Available: https://multilingual.com/noor-the-new-largest-nlp-model-for-the-arabic-language/

[11] H. Ramadan, “Naseej Launches its Innovative Arabic AI Language Model ‘Noon’ as an open-source initiative,” Naseej For Technology. Accessed: Oct. 07, 2024. [Online]. Available: https://naseej.com/news/2023/06/naseej-launches-its-innovative-arabic-ai-language-model-noon-as-an-open-source-initiative/

[12] “silma-ai/SILMA-9B-Instruct-v1.0 · Hugging Face.” Accessed: Oct. 07, 2024. [Online]. Available: https://huggingface.co/silma-ai/SILMA-9B-Instruct-v1.0

[13] “Watad.” Accessed: Aug. 30, 2024. [Online]. Available: https://watadenergy.com/mulhem/

[14] C. Malin, “Huawei Cloud reveals 100B Arabic LLM.” Accessed: Oct. 23, 2024. [Online]. Available: https://www.middleeastainews.com/p/huawei-reveals-100b-arabic-large-language-model

[15] W. Antoun, F. Baly, and H. Hajj, “AraELECTRA: Pre-Training Text Discriminators for Arabic Language Understanding,” in *Proceedings of the Sixth Arabic Natural Language Processing Workshop*, N. Habash, H. Bouamor, H. Hajj, W. Magdy, W. Zaghouani, F. Bougares, N. Tomeh, I. Abu Farha, and S. Touileb, Eds., Kyiv, Ukraine (Virtual): Association for Computational Linguistics, Apr. 2021, pp. 191–195. Accessed: Aug. 27, 2024. [Online]. Available: https://aclanthology.org/2021.wanlp-1.20

[16] R. AlYami and R. Al-Zaidy, “araRoperta,” in *Proceedings of the Seventh Arabic Natural Language Processing Workshop (WANLP)*, H. Bouamor, H. Al-Khalifa, K. Darwish, O. Rambow, F. Bougares, A. Abdelali, N. Tomeh, S. Khalifa, and W. Zaghouani, Eds., Abu Dhabi, United Arab Emirates (Hybrid): Association for Computational Linguistics, Dec. 2022, pp. 260–272. doi: 10.18653/v1/2022.wanlp-1.24.

[17] W. Lan, Y. Chen, W. Xu, and A. Ritter, “gigabert,” in *Proceedings of the 2020 Conference on Empirical Methods in Natural Language Processing (EMNLP)*, B. Webber, T. Cohn, Y. He, and Y. Liu, Eds., Online: Association for Computational Linguistics, Nov. 2020, pp. 4727–4734. doi: 10.18653/v1/2020.emnlp-main.382.

[18] A. Safaya, M. Abdullatif, and D. Yuret, “KUISAIL at SemEval-2020 Task 12: BERT-CNN for Offensive Speech Identification in Social Media,” in *Proceedings of the Fourteenth Workshop on Semantic Evaluation*, A. Herbelot, X. Zhu, A. Palmer, N. Schneider, J. May, and E. Shutova, Eds., Barcelona (online): International Committee for Computational Linguistics, Dec. 2020, pp. 2054–2059. doi: 10.18653/v1/2020.semeval-1.271.

[19] G. Inoue, B. Alhafni, N. Baimukan, H. Bouamor, and N. Habash, “camelbert,” in *Proceedings of the Sixth Arabic Natural Language Processing Workshop*, N. Habash, H. Bouamor, H. Hajj, W. Magdy, W. Zaghouani, F. Bougares, N. Tomeh, I. Abu Farha, and S. Touileb, Eds., Kyiv, Ukraine (Virtual): Association for Computational Linguistics, Apr. 2021, pp. 92–104. Accessed: Oct. 22, 2024. [Online]. Available: https://aclanthology.org/2021.wanlp-1.10

[20] M. Elgezouli, K. N. Elmadani, and M. Saeed, “SudaBERT: A Pre-trained Encoder Representation For Sudanese Arabic Dialect,” in *2020 International Conference on Computer, Control, Electrical, and Electronics Engineering (ICCCEEE)*, Feb. 2021, pp. 1–4. doi: 10.1109/ICCCEEE49695.2021.9429651.

[21] M. Al-qurishi, S. Alqaseemi, and R. Souissi, “AraLegal-BERT: A pretrained language model for Arabic Legal text,” in *Proceedings of the Natural Legal Language Processing Workshop 2022*, N. Aletras, I. Chalkidis, L. Barrett, C. Goan\textcommabelowtă, and D. Preo\textcommabelowtiuc-Pietro, Eds., Abu Dhabi, United Arab Emirates (Hybrid): Association for Computational Linguistics, Dec. 2022, pp. 338–344. doi: 10.18653/v1/2022.nllp-1.31.

[22] A. Ghaddar *et al.*, “jaber and saber,” in *Proceedings of the 2022 Conference on Empirical Methods in Natural Language Processing*, Y. Goldberg, Z. Kozareva, and Y. Zhang, Eds., Abu Dhabi, United Arab Emirates: Association for Computational Linguistics, Dec. 2022, pp. 3135–3151. doi: 10.18653/v1/2022.emnlp-main.205.

[23] A. Abdaoui, M. Berrimi, M. Oussalah, and A. Moussaoui, “DziriBERT: a Pre-trained Language Model for the Algerian Dialect,” Dec. 12, 2022, *arXiv*: arXiv:2109.12346. doi: 10.48550/arXiv.2109.12346.

[24] M. Kamal Eddine, N. Tomeh, N. Habash, J. Le Roux, and M. Vazirgiannis, “AraBART: a Pretrained Arabic Sequence-to-Sequence Model for Abstractive Summarization,” in *Proceedings of the Seventh Arabic Natural Language Processing Workshop (WANLP)*, H. Bouamor, H. Al-Khalifa, K. Darwish, O. Rambow, F. Bougares, A. Abdelali, N. Tomeh, S. Khalifa, and W. Zaghouani, Eds., Abu Dhabi, United Arab Emirates (Hybrid):

Association for Computational Linguistics, Dec. 2022, pp. 31–42. doi: 10.18653/v1/2022.wanlp-1.4.

[25] H. Haddad *et al.*, “TunBERT: Pretraining BERT for Tunisian Dialect Understanding,” *SN Comput. Sci.*, vol. 4, no. 2, p. 194, Feb. 2023, doi: 10.1007/s42979-022-01541-y.

[26] O. Moussaoui and Y. El Younoussi, “Pre-training Two BERT-Like Models for Moroccan Dialect: MorRoBERTa and MorrBERT,” *MENDEL*, vol. 29, pp. 55–61, Jun. 2023, doi: 10.13164/mendel.2023.1.055.

[27] F. Qarah, “SaudiBERT: A Large Language Model Pretrained on Saudi Dialect Corpora,” May 10, 2024, *arXiv*: arXiv:2405.06239. doi: 10.48550/arXiv.2405.06239.

[28] M. Ahmed, S. Alfasly, B. Wen, J. Addeen, M. Ahmed, and Y. Liu, “AlclaM: Arabic Dialect Language Model,” in *Proceedings of The Second Arabic Natural Language Processing Conference*, N. Habash, H. Bouamor, R. Eskander, N. Tomeh, I. Abu Farha, A. Abdelali, S. Touileb, I. Hamed, Y. Onaizan, B. Alhafni, W. Antoun, S. Khalifa, H. Haddad, I. Zitouni, B. AlKhamissi, R. Almatham, and K. Mrini, Eds., Bangkok, Thailand: Association for Computational Linguistics, Aug. 2024, pp. 153–159. doi: 10.18653/v1/2024.arabicnlp-1.14.

[29] F. Qarah, “EgyBERT: A Large Language Model Pretrained on Egyptian Dialect Corpora,” Aug. 06, 2024, *arXiv*: arXiv:2408.03524. doi: 10.48550/arXiv.2408.03524.

[30] F. Qarah, “AraPoemBERT: A Pretrained Language Model for Arabic Poetry Analysis,” Mar. 18, 2024, *arXiv*: arXiv:2403.12392. doi: 10.48550/arXiv.2403.12392.

[31] K. Gaanoun, A. M. Naira, A. Allak, and I. Benelallam, “DarijaBERT: a step forward in NLP for the written Moroccan dialect,” *Int. J. Data Sci. Anal.*, Jan. 2024, doi: 10.1007/s41060-023-00498-2.

[32] A. Dasari, “Understanding Encoder, Decoder, and Autoregressive Models in AI,” Medium. Accessed: Sep. 17, 2024. [Online]. Available: https://medium.com/@anusaid/understanding-encoder-decoder-and-autoregressive-models-in-ai-8da6ce9d4901

[33] P. P. Ray, “ChatGPT: A comprehensive review on background, applications, key challenges, bias, ethics, limitations and future scope,” *Internet Things Cyber-Phys. Syst.*, vol. 3, pp. 121–154, Jan. 2023, doi: 10.1016/j.iotcps.2023.04.003.

[34] W. Antoun, F. Baly, and H. Hajj, “AraGPT2: Pre-Trained Transformer for Arabic Language Generation,” in *Proceedings of the Sixth Arabic Natural Language Processing Workshop*, N. Habash, H. Bouamor, H. Hajj, W. Magdy, W. Zaghouani, F. Bougares, N. Tomeh, I. Abu Farha, and S. Touileb, Eds., Kyiv, Ukraine (Virtual): Association for Computational Linguistics, Apr. 2021, pp. 196–207. Accessed: Oct. 22, 2024. [Online]. Available: https://aclanthology.org/2021.wanlp-1.21

[35] E. M. Billah Nagoudi, M. Abdul-Mageed, A. Elmadany, A. Inciarte, and M. T. Islam Khondaker, “JASMINE: Arabic GPT Models for Few-Shot Learning,” in *Proceedings of the 2023 Conference on Empirical Methods in Natural Language Processing*, H. Bouamor, J. Pino, and K. Bali, Eds., Singapore: Association for Computational Linguistics, Dec. 2023, pp. 16721–16744. doi: 10.18653/v1/2023.emnlp-main.1040.

[36] N. Sengupta *et al.*, “Jais and Jais-chat: Arabic-Centric Foundation and Instruction-Tuned Open Generative Large Language Models,” Sep. 29, 2023, *arXiv*: arXiv:2308.16149. doi: 10.48550/arXiv.2308.16149.

[37] H. Touvron *et al.*, “LLaMA: Open and Efficient Foundation Language Models,” Feb. 27, 2023, *arXiv*: arXiv:2302.13971. doi: 10.48550/arXiv.2302.13971.

[38] H. Touvron *et al.*, “Llama 2: Open Foundation and Fine-Tuned Chat Models,” Jul. 19, 2023, *arXiv*: arXiv:2307.09288. doi: 10.48550/arXiv.2307.09288.

[39] H. Huang *et al.*, “AceGPT, Localizing Large Language Models in Arabic,” in *Proceedings of the 2024 Conference of the North American Chapter of the Association for Computational Linguistics: Human Language Technologies (Volume 1: Long Papers)*, K. Duh, H. Gomez, and S. Bethard, Eds., Mexico City, Mexico: Association for Computational Linguistics, Jun. 2024, pp. 8139–8163. doi: 10.18653/v1/2024.naacl-long.450.

[40] D. Groeneveld *et al.*, “OLMo: Accelerating the Science of Language Models,” in *Proceedings of the 62nd Annual Meeting of the Association for Computational Linguistics (Volume 1: Long Papers)*, L.-W. Ku, A. Martins, and V. Srikumar, Eds., Bangkok, Thailand: Association for Computational Linguistics, Aug. 2024, pp. 15789–15809. doi: 10.18653/v1/2024.acl-long.841.

[41] F. Team *et al.*, “Fanar: An Arabic-Centric Multimodal Generative AI Platform,” Jan. 18, 2025, *arXiv*: arXiv:2501.13944. doi: 10.48550/arXiv.2501.13944.

[42] G. Team *et al.*, “Gemma 2: Improving Open Language Models at a Practical Size,” Oct. 02, 2024, *arXiv*: arXiv:2408.00118. Accessed: Oct. 15, 2024. [Online]. Available: http://arxiv.org/abs/2408.00118

[43] G. Shang *et al.*, “Atlas-Chat: Adapting Large Language Models for Low-Resource Moroccan Arabic Dialect,” Sep. 26, 2024, *arXiv*: arXiv:2409.17912. Accessed: Oct. 02, 2024. [Online]. Available: http://arxiv.org/abs/2409.17912

[44] “Stable LM 2 1.6B Technical Report.” Accessed: Feb. 16, 2025. [Online]. Available: https://arxiv.org/html/2402.17834v1

[45] Z. Alyafeai *et al.*, “Arabic Stable LM: Adapting Stable LM 2 1.6B to Arabic,” Dec. 05, 2024, *arXiv*: arXiv:2412.04277. doi: 10.48550/arXiv.2412.04277.

[46] A. Alghamdi *et al.*, “AraMUS: Pushing the Limits of Data and Model Scale for Arabic Natural Language Processing,” in *Findings of the Association for Computational Linguistics: ACL 2023*, A. Rogers, J. Boyd-Graber, and N. Okazaki, Eds., Toronto, Canada: Association for Computational Linguistics, Jul. 2023, pp. 2883–2894. doi: 10.18653/v1/2023.findings-acl.181.

[47] E. M. B. Nagoudi, A. Elmadany, and M. Abdul-Mageed, “AraT5: Text-to-Text Transformers for Arabic Language Generation,” in *Proceedings of the 60th Annual Meeting of the Association for Computational Linguistics (Volume 1: Long Papers)*, S. Muresan, P. Nakov, and A. Villavicencio, Eds., Dublin, Ireland: Association for Computational Linguistics, May 2022, pp. 628–647. doi: 10.18653/v1/2022.acl-long.47.

[48] Y. Adel *et al.*, “AraQA: An Arabic Generative Question-Answering Model for Authentic Religious Text,” in *2023 11th International Japan-Africa Conference on Electronics, Communications, and Computations (JAC-ECC)*, Dec. 2023, pp. 235–239. doi: 10.1109/JAC-ECC61002.2023.10479645.

[49] A. Koubaa, A. Ammar, L. Ghouti, O. Najar, and S. Sibaee, “ArabianGPT: Native Arabic GPT-based Large Language Model,” Feb. 26, 2024, *arXiv*: arXiv:2402.15313. doi: 10.48550/arXiv.2402.15313.

[50] A. El-Shangiti, F. Alwajih, and M. Abdul-Mageed, “Arabic Automatic Story Generation with Large Language Models,” in *Proceedings of The Second Arabic Natural Language Processing Conference*, N. Habash, H. Bouamor, R. Eskander, N. Tomeh, I. Abu Farha, A. Abdelali, S. Touileb, I. Hamed, Y. Onaizan, B. Alhafni, W. Antoun, S. Khalifa, H. Haddad, I. Zitouni, B. AlKhamissi, R. Almatham, and K. Mrini, Eds., Bangkok, Thailand: Association for Computational Linguistics, Aug. 2024, pp. 140–152. Accessed: Sep. 06, 2024. [Online]. Available: https://aclanthology.org/2024.arabicnlp-1.13

[51] M. S. Bari *et al.*, “ALLaM: Large Language Models for Arabic and English,” Jul. 22, 2024, *arXiv*: arXiv:2407.15390. Accessed: Aug. 28, 2024. [Online]. Available: http://arxiv.org/abs/2407.15390

[52] J. Zhu *et al.*, “Second Language (Arabic) Acquisition of LLMs via Progressive Vocabulary Expansion,” Dec. 16, 2024, *arXiv*: arXiv:2412.12310. doi: 10.48550/arXiv.2412.12310.

[53] “Modern Standard Arabic,” *Wikipedia*. Aug. 15, 2024. Accessed: Aug. 29, 2024. [Online]. Available: https://en.wikipedia.org/w/index.php?title=Modern_Standard_Arabic&oldid=1240502833

[54] H. Bouamor, N. Habash, and K. Oflazer, “A Multidialectal Parallel Corpus of Arabic”.

[55] W. Zaghouani and A. Charfi, “Arap-Tweet: A Large Multi-Dialect Twitter Corpus for Gender, Age and Language Variety Identification,” in *Proceedings of the Eleventh International Conference on Language Resources and Evaluation (LREC 2018)*, N. Calzolari, K. Choukri, C. Cieri, T. Declerck, S. Goggi, K. Hasida, H. Isahara, B. Maegaard, J. Mariani, H. Mazo, A. Moreno, J. Odijk, S. Piperidis, and T. Tokunaga, Eds., Miyazaki, Japan: European Language Resources Association (ELRA), May 2018. Accessed: Sep. 16, 2024. [Online]. Available: https://aclanthology.org/L18-1111

[56] A. Conneau *et al.*, “Unsupervised Cross-lingual Representation Learning at Scale,” in *Proceedings of the 58th Annual Meeting of the Association for Computational Linguistics*, Online: Association for Computational Linguistics, 2020. doi: 10.18653/v1/2020.acl-main.747.

[57] G. Inoue, B. Alhafni, N. Baimukan, H. Bouamor, and N. Habash, “The Interplay of Variant, Size, and Task Type in Arabic Pre-trained Language Models,” Sep. 04, 2021, *arXiv*: arXiv:2103.06678. doi: 10.48550/arXiv.2103.06678.

[58] B. Talafha *et al.*, “Multi-dialect Arabic BERT for Country-level Dialect Identification,” in *Proceedings of the Fifth Arabic Natural Language Processing Workshop*, I. Zitouni, M. Abdul-Mageed, H. Bouamor, F. Bougares, M. El-Haj, N. Tomeh, and W. Zaghouani, Eds., Barcelona, Spain (Online): Association for Computational Linguistics, Dec. 2020, pp. 111–118. Accessed: Sep. 20, 2024. [Online]. Available: https://aclanthology.org/2020.wanlp-1.10

[59] Y. Liu *et al.*, “RoBERTa: A Robustly Optimized BERT Pretraining Approach,” Jul. 26, 2019, *arXiv*: arXiv:1907.11692. doi: 10.48550/arXiv.1907.11692.

[60] F. Feng, Y. Yang, D. Cer, N. Arivazhagan, and W. Wang, “Language-agnostic BERT Sentence Embedding,” in *Proceedings of the 60th Annual Meeting of the Association for Computational Linguistics (Volume 1: Long Papers)*, S. Muresan, P. Nakov, and A. Villavicencio, Eds., Dublin, Ireland: Association for Computational Linguistics, May 2022, pp. 878–891. doi: 10.18653/v1/2022.acl-long.62.

[61] T. Kudo and J. Richardson, “SentencePiece: A simple and language independent subword tokenizer and detokenizer for Neural Text Processing,” in *Proceedings of the 2018 Conference on Empirical Methods in Natural Language Processing: System Demonstrations*, E. Blanco and W. Lu, Eds., Brussels, Belgium: Association for Computational Linguistics, Nov. 2018, pp. 66–71. doi: 10.18653/v1/D18-2012.

[62] R. Zellers *et al.*, “Defending against neural fake news,” in *Proceedings of the 33rd International Conference on Neural Information Processing Systems*, Red Hook, NY, USA: Curran Associates Inc., 2019, pp. 9054–9065.

[63] O. Shliazhko, A. Fenogenova, M. Tikhonova, A. Kozlova, V. Mikhailov, and T. Shavrina, “mGPT: Few-Shot Learners Go Multilingual,” *Trans. Assoc. Comput. Linguist.*, vol. 12, pp. 58–79, 2024, doi: 10.1162/tacl_a_00633.

[64] S. Rothe, S. Narayan, and A. Severyn, “Leveraging Pre-trained Checkpoints for Sequence Generation Tasks,” *Trans. Assoc. Comput. Linguist.*, vol. 8, pp. 264–280, Dec. 2020, doi: 10.1162/tacl_a_00313.

[65] Y. Liu *et al.*, “Multilingual Denoising Pre-training for Neural Machine Translation,” *Trans. Assoc. Comput. Linguist.*, vol. 8, pp. 726–742, Dec. 2020, doi: 10.1162/tacl_a_00343.

[66] L. Xue *et al.*, “mT5: A massively multilingual pre-trained text-to-text transformer,” Mar. 11, 2021, *arXiv*: arXiv:2010.11934. Accessed: Oct. 04, 2024. [Online]. Available: http://arxiv.org/abs/2010.11934

[67] A. Vaswani *et al.*, “Attention Is All You Need,” Aug. 01, 2023, *arXiv*: arXiv:1706.03762. Accessed: Oct. 04, 2024. [Online]. Available: http://arxiv.org/abs/1706.03762

[68] A. Üstün *et al.*, “Aya Model: An Instruction Finetuned Open-Access Multilingual Language Model,” Feb. 12, 2024, *arXiv*: arXiv:2402.07827. Accessed: Oct. 12, 2024. [Online]. Available: http://arxiv.org/abs/2402.07827

[69] “Command R: RAG at Production Scale,” Cohere. Accessed: Oct. 12, 2024. [Online]. Available: https://cohere.com/blog/command-r

[70] “BLOOM.” Accessed: Oct. 12, 2024. [Online]. Available: https://bigscience.huggingface.co/blog/bloom

[71] “Qwen (Qwen).” Accessed: Oct. 12, 2024. [Online]. Available: https://huggingface.co/Qwen

[72] “Cohere | The leading AI platform for enterprise.” Accessed: Oct. 12, 2024. [Online]. Available: https://cohere.com/

[73] “Alibaba Cloud: Reliable Secure Cloud Solutions to Empower Your Global Business.” Accessed: Oct. 12, 2024. [Online]. Available: https://www.alibabacloud.com/en?_p_lc=7

[74] “Introducing Gemini: our largest and most capable AI model,” Google. Accessed: Oct. 12, 2024. [Online]. Available: https://blog.google/technology/ai/google-gemini-ai/

[75] J.-C. Burgelman *et al.*, “Open Science, Open Data, and Open Scholarship: European Policies to Make Science Fit for the Twenty-First Century,” *Front. Big Data*, vol. 2, Dec. 2019, doi: 10.3389/fdata.2019.00043.

[76] A. Birhane, A. Kasirzadeh, D. Leslie, and S. Wachter, “Science in the age of large language models,” *Nat. Rev. Phys.*, vol. 5, no. 5, pp. 277–280, Apr. 2023, doi: 10.1038/s42254-023-00581-4.

[77] D. Gray Widder, S. West, and M. Whittaker, “Open (For Business): Big Tech, Concentrated Power, and the Political Economy of Open AI,” *SSRN Electron. J.*, 2023, doi: 10.2139/ssrn.4543807.

[78] A. Liesenfeld and M. Dingemanse, “Rethinking open source generative AI: open washing and the EU AI Act,” in *The 2024 ACM Conference on Fairness, Accountability, and Transparency*, Rio de Janeiro Brazil: ACM, Jun. 2024, pp. 1774–1787. doi: 10.1145/3630106.3659005.

[79] “EU AI Act: first regulation on artificial intelligence,” Topics | European Parliament. Accessed: Oct. 12, 2024. [Online]. Available: https://www.europarl.europa.eu/topics/en/article/20230601STO93804/eu-ai-act-first-regulation-on-artificial-intelligence

[80] M. White, I. Haddad, C. Osborne, X.-Y. L. Yanglet, A. Abdelmonsef, and S. Varghese, “The Model Openness Framework: Promoting Completeness and Openness for Reproducibility, Transparency, and Usability in Artificial Intelligence,” Oct. 02, 2024, *arXiv*: arXiv:2403.13784. Accessed: Oct. 16, 2024. [Online]. Available: http://arxiv.org/abs/2403.13784

[81] R. Kitchin, *The Data Revolution: Big Data, Open Data, Data Infrastructures and Their Consequences*. SAGE, 2014.

[82] C. Osborne, J. Ding, and H. R. Kirk, “The AI Community Building the Future? A Quantitative Analysis of Development Activity on Hugging Face Hub,” Jun. 05, 2024, *arXiv*: arXiv:2405.13058. Accessed: Oct. 16, 2024. [Online]. Available: http://arxiv.org/abs/2405.13058

[83] M. Mitchell *et al.*, “Model Cards for Model Reporting,” Jan. 14, 2019, *arXiv*: arXiv:1810.03993. Accessed: Oct. 16, 2024. [Online]. Available: http://arxiv.org/abs/1810.03993